\documentclass[10pt,twocolumn,letterpaper]{article}
\usepackage{preprint-layout}
\usepackage{amsmath,amssymb}
\usepackage{bbm,algorithm,algpseudocode}
\usepackage{graphicx}
\graphicspath{{figures/}}
\usepackage{booktabs,multirow,tabularx,array,placeins,longtable,float}
\usepackage{xcolor,microtype,enumitem,url}
\definecolor{linkblue}{rgb}{0.21,0.49,0.74}
\usepackage[breaklinks,colorlinks,allcolors=linkblue]{hyperref}
\pdftrailerid{}
\hypersetup{pdftitle={Seeing and Solving Are Not Enough for Vision-Language Models},pdfauthor={Ziheng Wang, Shiying Dong, Mingxuan Xie, Yilin LIU, Dayan Wu, Yang Li, Pengwen Dai},pdfcreator={},pdfproducer={}}
\makeatletter
\providecommand{\theHALG@line}{}
\renewcommand{\theHALG@line}{supp.\arabic{algorithm}.\arabic{ALG@line}}
\makeatother

\newcommand{\cfs}{\mathrm{CFS}}
\newcommand{\cfr}{\mathrm{CFR}}
\newcommand{\ind}{\mathbbm{1}}
\newcommand{\concat}{\mathrm{Concat}}
\newcommand{\Fset}{\mathcal F}
\newcommand{\Fext}{\mathcal F_{\mathrm{ext}}}
\newcommand{\Fsol}{\mathcal F_{\mathrm{sol}}}
\newcommand{\Fcmp}{\mathcal F_{\mathrm{cmp}}}
\newcommand{\Jset}{\mathcal J}
\newcommand{\Gset}{\mathcal G}
\newcommand{\ms}[2]{#1\pm#2}
\newcommand{\rD}{\mathrm D}
\newcommand{\rC}{\mathrm C}
\newcommand{\tablecite}[1]{~\cite{#1}}
\title{Seeing and Solving Are Not Enough for Vision-Language Models}
\author{%
  Ziheng Wang$^{1}$ \quad Shiying Dong$^{1}$ \quad Mingxuan Xie$^{2}$ \quad Yilin LIU$^{3}$\\[2pt]
  Dayan Wu$^{4}$ \quad Yang Li$^{5}$ \quad Pengwen Dai$^{1,*}$\\[5pt]
  {\small $^{1}$Sun Yat-sen University \quad $^{2}$Zhejiang University}\\
  {\small $^{3}$The Hong Kong University of Science and Technology}\\
  {\small $^{4}$Institute of Information Engineering \quad $^{5}$Hunan University}\\[3pt]
  {\small $^{*}$Corresponding author: \href{mailto:daipw@mail.sysu.edu.cn}{\texttt{daipw@mail.sysu.edu.cn}}}%
}

\begin{document}
\twocolumn[{%
\renewcommand\twocolumn[1][]{#1}%
\maketitle
\begin{center}
\includegraphics[width=0.92\textwidth]{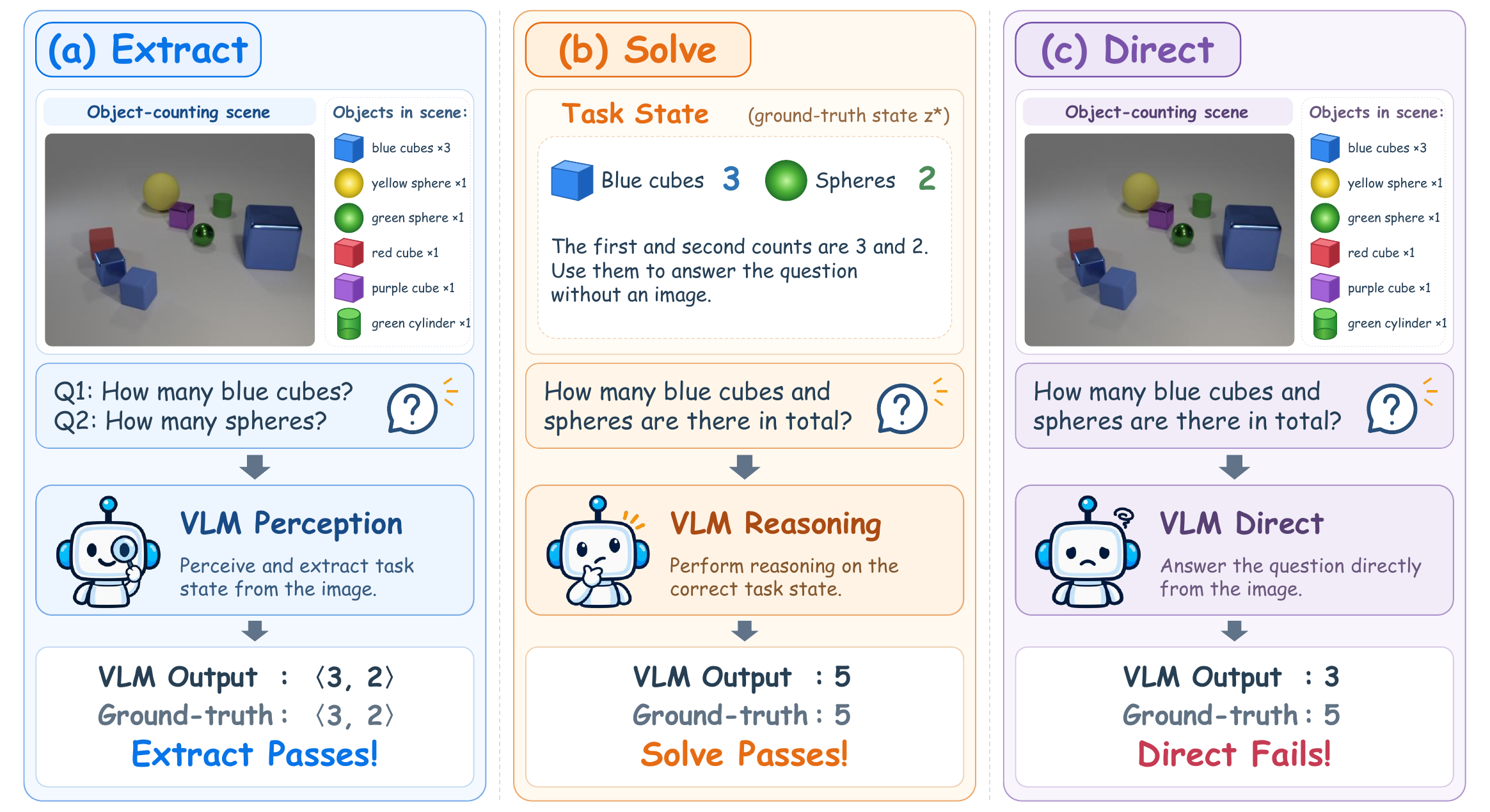}
\captionsetup{hypcap=false}
\captionof{figure}{An illustration of a composition failure. On the same visual question, the model recovers the task state $z^\star$ from the image (Extract) and computes the correct answer from $z^\star$ without the image (Solve), yet fails to answer directly from the image (Direct).}
\captionsetup{hypcap=true}
\label{fig:teaser}
\end{center}
}]

\begin{abstract}
Vision-language models (VLMs) answer visual questions by extracting visual information from an image and reasoning over it. However, final-answer accuracy cannot tell whether an error arises from extraction, solving, or combining the two. To investigate this issue, we propose \textbf{StateLens}, which reverses conventional question construction by first verifying the visual information sufficient to determine the answer and then generating questions from it. On each constructed question, StateLens evaluates extraction, solving, and direct answering separately, and further decouples chain-of-thought (CoT) supervision into visual facts and derivations. Across five VLMs, \emph{composition failures}, where extraction and solving both succeed yet direct answering fails, account for 40.2--75.6\% of direct-answering errors on StateLens ChartQA and remain frequent across multi-structure and scene tasks. Moreover, writing visual facts before the answer repairs 90.2--99.6\% of these failures, explicit derivations mainly benefit multi-step arithmetic, and correct answers can still rely on wrong visual facts. Motivated by these findings, we propose \textbf{Parallax}, which pairs direct answering with a reasoning challenger and learns when to trust reasoning through source-bound answer verification against the image. On the original ChartQA benchmark, Parallax improves accuracy from 86.68\% to 91.44\%, outperforming CoT prompting, native thinking, and larger VLMs. On the StateLens ChartQA benchmark, it further recovers 98.8\% of repairable composition failures.
\end{abstract}

\section{Introduction}
\label{sec:intro}

Visual question answering (VQA) requires both detailed image understanding and reasoning~\cite{antol2015vqa,johnson2017clevr,hudson2019gqa,masry2022chartqa}, yet an incorrect final answer alone does not reveal which step failed. Consider a chart question asking for the sum of two values. A vision-language model (VLM)~\cite{li2023blip2,liu2023visualinstruction,bai2025qwen25vl} may accurately recover both values in isolation and compute their sum when the values are given, yet still fail when asked to answer directly from the image. We call this pattern a \emph{composition failure}, where visual extraction and problem solving both succeed on the same question when tested separately, yet direct multimodal answering fails.

Gaps between component abilities and final answers have been observed before. Press et al.~\cite{press2023measuring} showed that language models can solve decomposed subquestions yet fail to combine them on the original query, calling this the compositionality gap. For multimodal questions, however, diagnosing such a failure also requires knowing whether the model recovered the visual information needed for the question in the first place. In VLMs, Prism~\cite{qiao2024prism} and DISSECT~\cite{kukreja2026dissect} separate visual extraction from reasoning by verbalizing the image before reasoning over the description, while recent suites probe component skills via step-level subproblems or atomic visual queries~\cite{qiao2025wemath,chung2025mathlens}. Because free-form descriptions cannot be scored exactly and atomic probes test only isolated details, it remains unclear for a failed question whether all required visual information was recovered. Meanwhile, chain-of-thought (CoT) supervision writes intermediate steps before the answer~\cite{wei2022chain,zhang2024multimodal,xu2024llavacot}, yet it entangles visual facts with calculations. Moreover, it yields variable gains across tasks~\cite{jin2026looklight,bi2026reasoningseeing,cai2026lookthink,zhang2024tinychart} and does not necessarily state faithful facts~\cite{jacovi2020faithfulness,lanham2023faithfulness}. Together, these limitations leave open whether extraction and solving both succeed on the same failed question, and which part of intermediate supervision repairs such composition failures.

We make this distinction measurable through a \emph{task state}, which represents the visual information sufficient to solve a question in an exactly scorable form. Building on ChartQA tables and CLEVR programs~\cite{masry2022chartqa,johnson2017clevr}, we propose \textbf{StateLens}, which reverses question construction by first selecting doubly verified visual facts and then generating the task state and question. On each question, we evaluate the same model through three interfaces: \emph{Extract} recovers the task state from the image and question, \emph{Solve} answers from the ground-truth state without the image, and \emph{Direct} answers from the image and question directly. As illustrated in Figure~\ref{fig:teaser}, a composition failure occurs when Extract and Solve both succeed on the same question yet Direct fails. To test what repairs these failures, StateLens reuses the verified states to fine-tune LoRA adapters~\cite{hu2022lora} in the language layers of the frozen VLM across six supervision targets. Across five VLMs, composition failures account for 40.2--75.6\% of Direct errors on StateLens ChartQA and remain widespread on CLEVR and multi-structure tasks. Decoupling supervision further reveals that writing visual facts before the answer repairs 90.2--99.6\% of composition failures, explicit derivations mainly benefit multi-step arithmetic while compact states match CoT elsewhere with half the tokens, and correct answers can still rely on wrong visual facts.

These findings show that while explicit reasoning can repair composition failures, its generated facts and answers are not always reliable. In fact, applying an unguarded reasoning route to every question on the original ChartQA benchmark lowers accuracy by overturning many correct direct answers. To address this on natural questions without annotated states, we propose \textbf{Parallax} to learn when to trust reasoning over direct answering. Parallax treats the direct and reasoning routes as two complementary viewpoints and triggers verification only when their valid answers disagree. Conditioned on the image and question, a LoRA selector compares the two candidate answers, using a source header to mark the reasoning challenger while withholding its unverified draft. On the original ChartQA benchmark, Parallax achieves the highest accuracy among the compared methods, surpassing CoT prompting, native thinking, chart-specific models, and larger VLMs. It also recovers nearly all repairable composition failures on StateLens ChartQA without retraining.

Our contributions are summarized as follows:
\begin{itemize}[leftmargin=*,topsep=1pt,itemsep=1pt]
\item We propose \textbf{StateLens}, a question-level diagnostic framework that uses doubly verified, exactly scorable task states to evaluate visual extraction, problem solving, and direct answering on the same question across five VLMs.
\item Using \textbf{StateLens} to decouple intermediate supervision, we show that pre-answer visual facts repair most composition failures, derivations mainly benefit multi-step arithmetic, and correct answers do not guarantee accurate visual facts.
\item We introduce \textbf{Parallax} for source-bound answer verification on natural questions, improving the direct-answering accuracy on ChartQA by 4.76 points on Qwen3.5 with state-of-the-art performance and 21.20 points on Molmo.
\end{itemize}

\section{Related Work}
\label{sec:related}

\noindent\textbf{Visual question answering benchmarks.}
VQA benchmarks test spatial relations and counting in scenes~\cite{antol2015vqa,johnson2017clevr,hudson2019gqa} as well as value reading and arithmetic on charts~\cite{methani2020plotqa,masry2022chartqa}. Recent suites broaden chart styles, multi-chart synthesis, and misleading plots~\cite{zhu2025multichartqa,masry2025chartqapro,chen2026chartr,lalai2026misleading} or probe language priors, binding, and scene fragmentation~\cite{ma2023crepe,hsieh2023sugarcrepe,hua2024mmcomposition,teja2026layeredvqa}. Although several datasets provide source tables or scene annotations, standard evaluations score only final answers or caption matches without verifying the visual facts needed per question. StateLens uses ChartQA and CLEVR annotations to construct questions paired with verified task states.

\noindent\textbf{Component-level diagnosis.}
Press et al.~\cite{press2023measuring} measure the compositionality gap in language models via subquestion and multi-hop accuracies. In VLMs, Prism and DISSECT separate perception and reasoning via descriptions or modified inputs~\cite{qiao2024prism,kukreja2026dissect}, We-Math decomposes visual math into subproblems~\cite{qiao2025wemath}, VisRes Bench and ChartProbe use targeted probes~\cite{tortei2026visres,khoshnoodi2026chartprobe}, and MathLens builds atomic geometry probes~\cite{chung2025mathlens}. Recent studies also analyze single-pass composition~\cite{wang2026compose}, visual attention and operation pathways~\cite{liu2026seeing,gupta2026visionop}, false premises and process priors~\cite{liu2025unveiling,dai2026seeingnotenough}, and primitive integration load~\cite{li2025compabench,sbrolli2026autocomp,tian2026crosscheck,gandhi2026compass}. However, descriptions, isolated probes, and marginal averages cannot verify whether a model both extracts the visual information sufficient to answer a failed question and solves from it on the same instance. StateLens formalizes this sufficient visual information as a question-level task state.

\noindent\textbf{Chain-of-thought and intermediate supervision.}
Intermediate outputs are widely used to support reasoning, from scratchpads and CoT in language models~\cite{nye2021scratchpads,wei2022chain} to multimodal rationales~\cite{zhang2024multimodal,xu2024llavacot,wu2026groundedcot}, visual programs~\cite{gupta2023visual,suris2023vipergpt}, chart tables or traces~\cite{liu2023deplot,zhang2024tinychart,hu2024vpd,chartmoe2025}, and decoupled post-training~\cite{lee2026visdot,chen2026mathflow,li2025visionsr1,chen2026cogflow,wu2026seeingthinking}. However, CoT gains vary across tasks~\cite{jin2026looklight,bi2026reasoningseeing}, longer traces increase decoding cost~\cite{cai2026lookthink}, and generated rationales need not state faithful facts~\cite{jacovi2020faithfulness,lanham2023faithfulness}. Because free-form CoT entangles visual extraction with calculations, it remains unclear which part repairs composition failures. StateLens decouples target content, position, form, and derivations under one shared recipe and audits every written fact against the verified task state.

\noindent\textbf{Answer selection and verification.}
Given multiple candidates, self-consistency votes over sampled paths~\cite{wang2023selfconsistency}, LLM-Blender ranks and fuses candidates~\cite{jiang2023llmblender}, and LLM judges score rationales while showing position bias~\cite{zheng2023judging}. In VLMs, LLaVA-Critic learns pairwise image-conditioned judging~\cite{xiong2025llavacritic}, and MM-Verify scores full reasoning traces~\cite{sun2025mmverify}. However, symmetric selection ignores that an unguarded reasoning route often breaks correct direct answers, while judging rationales relies on unverified facts. In contrast, Parallax anchors on the direct answer, gates on valid disagreements, and compares final answers across both orders with source awareness and the draft withheld.

\section{Method}
\label{sec:method}

\begin{figure*}[!t]
\centering
\includegraphics[width=0.92\textwidth]{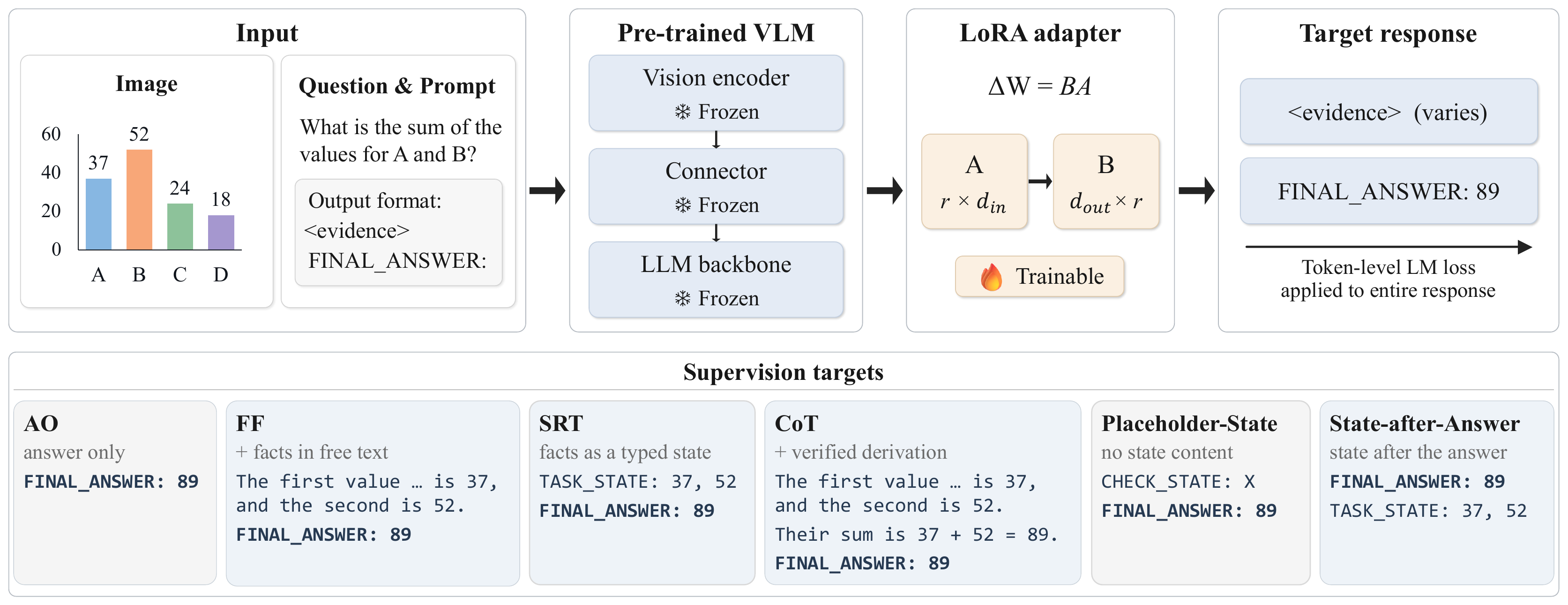}
\caption{Training framework and supervision targets in StateLens. Separate LoRA adapters are trained in the language layers of a frozen VLM under matched optimization settings, varying the content, position, form, and derivation of the target response.}
\label{fig:conditions}
\end{figure*}

\subsection{Task State and Composition Failures}
\label{sec:problem}

\noindent\textbf{Task state formulation.}
Given an image $I$ and a question $q$, a VLM $M_\theta$ with parameters $\theta$ predicts an answer $\hat y = M_\theta(I,q)$. Its accuracy over $N$ questions is $\mathrm{Acc} = \frac{1}{N}\sum_{i=1}^{N}\ind[\kappa(\hat y_i)=\kappa(y_i^\star)]$, where $\ind[\cdot]$ is the indicator function and $\kappa(\cdot)$ canonicalizes responses, mapping malformed outputs to an unmatched null symbol. However, final-answer accuracy measures only end-to-end correctness and cannot distinguish among errors in visual extraction, downstream reasoning, and their composition. To evaluate extraction and reasoning separately, we consider the structured source $W$ from which $I$ is generated, such as the data table of a chart or the scene graph of a 3D scene. We define the \emph{task state} $z^\star\in\mathcal Z_q$ as the question-relevant visual information in $W$ sufficient to determine the answer, where the state space $\mathcal Z_q$ fixes the structure required by $q$, such as an ordered pair, a table, or a set. The answer is then factorized as
\begin{equation}
I \sim p(I\mid W),\qquad y^\star = f(q, W) = \phi\big(q, \underbrace{g_q(W)}_{z^\star}\big),
\label{eq:task_state}
\end{equation}
where $p(I\mid W)$ denotes how images are generated from $W$ with possibly different chart types, styles, or viewpoints. The function $g_q$ reads the facts required by $q$ from $W$ without applying the operation in $q$, and $\phi$ then applies this operation to $z^\star$. For example, if $q$ asks for the sum of two bars, $z^\star$ is the pair of bar values for any drawing of the chart, and $\phi$ adds the two values. Since $y^\star$ depends on $W$ only through $z^\star$, it can be computed from $q$ and $z^\star$ without the image.

\noindent\textbf{Question-level composition failures.}
Using the ground-truth task state $z_i^\star$, we evaluate the same model $M_\theta$ on each instance $x_i=(I_i,q_i,z_i^\star,y_i^\star)$ through three interfaces, namely \emph{Extract}, \emph{Solve}, and \emph{Direct}, none of which requires an explicit reasoning trace. Their predictions and correctness indicators $E_i,S_i,D_i\in\{0,1\}$ are
\begin{equation}
\begin{aligned}
\hat z_i &= M_\theta(I_i,q_i,\pi^{\mathrm E}), & E_i &= \ind\big[\kappa(\hat z_i)=\kappa(z_i^\star)\big],\\
\hat y_i^{\mathrm S} &= M_\theta(q_i,z_i^\star,\pi^{\mathrm S}), & S_i &= \ind\big[\kappa(\hat y_i^{\mathrm S})=\kappa(y_i^\star)\big],\\
\hat y_i^{\rD} &= M_\theta(I_i,q_i,\pi^{\rD}), & D_i &= \ind\big[\kappa(\hat y_i^{\rD})=\kappa(y_i^\star)\big].
\end{aligned}
\label{eq:interfaces}
\end{equation}
Here, $\pi^{\mathrm E}$, $\pi^{\mathrm S}$, and $\pi^{\rD}$ prompt the model to extract the task state from $(I_i,q_i)$, solve the question from $(q_i,z_i^\star)$ without the image, and answer directly from $(I_i,q_i)$, respectively. Question $i$ is a \emph{composition failure} if $E_i=S_i=1$ and $D_i=0$, meaning that the model extracts and solves correctly in separate calls yet fails to answer directly.

The marginal accuracies $\bar E$, $\bar S$, and $\bar D$ do not determine the proportion $\rho$ of composition failures among the $N$ questions and only constrain it through the Fr\'echet bounds, as derived in the supplementary material. We therefore partition the Direct errors $\Fset=\{i:D_i=0\}$ at the question level into extraction failures $\Fext$ with $E_i=0$, solving failures $\Fsol$ with $E_i=1$ and $S_i=0$, and composition failures $\Fcmp$ with $E_i=S_i=1$. With $\Jset=\{i:E_i=S_i=1\}$ denoting the jointly solvable questions, the composition failure share (CFS) and rate (CFR) are defined as
\begin{equation}
\cfs = \frac{|\Fcmp|}{|\Fset|},\qquad \cfr = \frac{|\Fcmp|}{|\Jset|},
\label{eq:metrics}
\end{equation}
which measure the share of Direct errors that are composition failures and the Direct error rate on $\Jset$~\cite{press2023measuring}.

\subsection{StateLens: Decoupling Multimodal Reasoning}
\label{sec:statelens}

Standard VQA datasets typically provide final answers without verified intermediate task states, while conventional CoT mixes visual facts with subsequent derivations. StateLens constructs questions in reverse from doubly verified visual facts and uses the resulting task states to identify composition failures and decouple intermediate supervision.

\noindent\textbf{Reverse question construction.}
Constructing questions before identifying their required visual facts can leave task states incomplete or ambiguous. StateLens reverses this order by verifying visual facts before generating questions. Given an image $I$ and its structured dataset annotations $\mathcal A$, we first select doubly verified visual records $\mathcal A^\star$ and then generate the question $q$, task state $z^\star$, and answer $y^\star$ as
\begin{equation}
\begin{aligned}
\mathcal A^\star &= \sigma\big(\{a\subseteq\mathcal A : V(a)=1\}\big),\quad
q = \psi_q\big(\ell(\mathcal A^\star), o\big),\\
z^\star &= \psi_z\big(v(\mathcal A^\star)\big),\qquad\qquad\quad\;\;
y^\star = \phi_o(z^\star).
\end{aligned}
\label{eq:construction}
\end{equation}
Here, $a\subseteq\mathcal A$ denotes a candidate subset of annotated records, such as two table cells of a chart, and $V(a)\in\{0,1\}$ indicates whether its values agree across two independent annotation sources, such as the source table and the chart annotation. The selection function $\sigma$ deterministically chooses one verified subset $\mathcal A^\star$, from which $\ell(\cdot)$ and $v(\cdot)$ extract labels and values, respectively. Given a reasoning operation $o$ such as a sum, $\psi_q$ generates the question from the selected labels, $\psi_z$ formats the corresponding values into the task state, and $\phi_o$ applies $o$ to these values to compute the answer. In Figure~\ref{fig:conditions}, for example, $\mathcal A^\star$ consists of the verified records of bars A and B, with $z^\star=(37,52)$ and $y^\star=89$. Fixing the verified visual facts before question generation ensures that $z^\star$ is sufficient to determine $y^\star$ and that Extract and Direct refer to the same visual facts. The same facts are also written in prose as $e^\star$, and executing $\phi_o$ step by step gives the derivation $d^\star$. The construction algorithms for each benchmark are provided in the supplementary material.

\begin{figure*}[!t]
\centering
\includegraphics[width=0.92\textwidth]{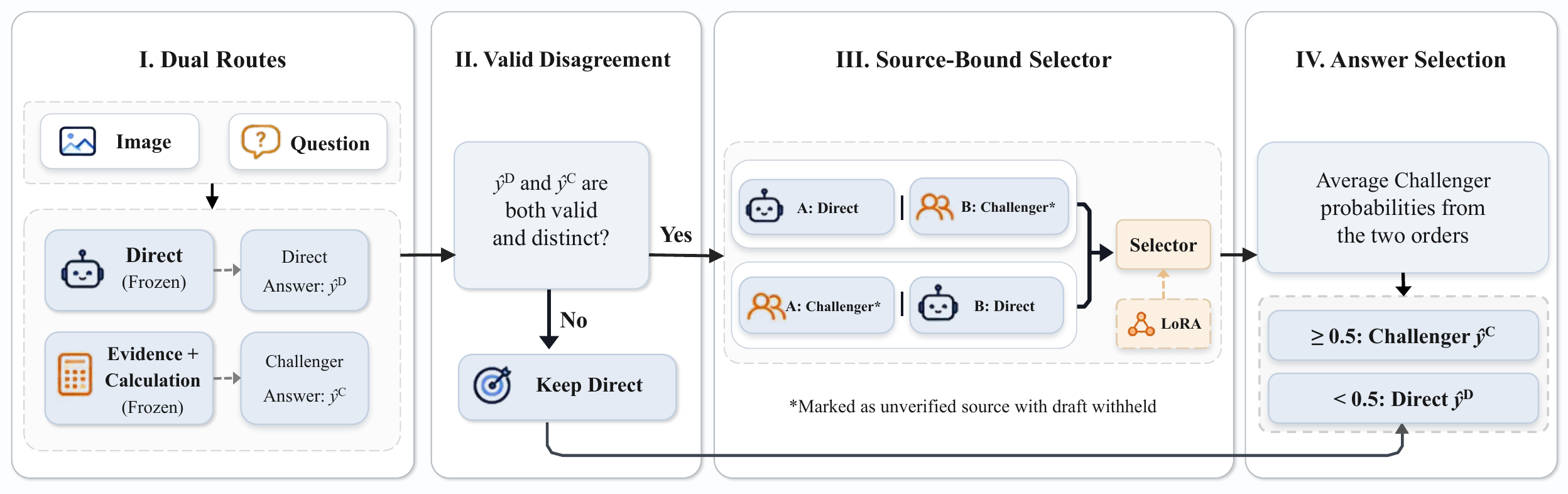}
\caption{Overview of Parallax. Given an image and question, frozen Direct and Challenger routes output candidate answers $\hat y^{\rD}$ and $\hat y^{\rC}$. When both answers are valid and distinct, a LoRA selector compares them in both option orders with an unverified-source header attached to the Challenger answer and its draft withheld, switching to $\hat y^{\rC}$ only when its average probability reaches $0.5$.}
\label{fig:parallax}
\end{figure*}

\noindent\textbf{Supervision variants and conditional LoRA training.}
Because standard CoT combines visual facts $e^\star$ with step-by-step derivations $d^\star$, final-answer accuracy alone cannot distinguish their respective contributions to repairing composition failures. As illustrated in Figure~\ref{fig:conditions}, StateLens compares six supervision targets $R_c$ sharing $y^\star$, defined as
\[
\begin{array}{@{}r@{}l@{\quad}r@{}l@{}}
R_{\mathrm{AO}} &{}= y^\star, & R_{\mathrm{PS}} &{}= \concat(z_\varnothing, y^\star),\\
R_{\mathrm{SA}} &{}= \concat(y^\star, z^\star), & R_{\mathrm{SRT}} &{}= \concat(z^\star, y^\star),\\
R_{\mathrm{FF}} &{}= \concat(e^\star, y^\star), & R_{\mathrm{CoT}} &{}= \concat(e^\star, d^\star, y^\star),
\end{array}
\]
where $\concat$ denotes the ordered concatenation of text fields and $z_\varnothing$ is a fixed placeholder that carries no visual information. \emph{Answer-Only} (AO) supervises $y^\star$ alone. \emph{Placeholder-State} (PS) prepends $z_\varnothing$ to control for the additional output structure. \emph{State-after-Answer} (SA) places $z^\star$ after $y^\star$, supervising the state without conditioning the answer on it. \emph{State Realization Tuning} (SRT) places the structured state $z^\star$ before $y^\star$, \emph{Fact-First} (FF) states the same visual facts in prose as $e^\star$, and \emph{Chain-of-Thought} (CoT)~\cite{wei2022chain} further adds the verified derivation $d^\star$ before the answer. These comparisons assess informative pre-answer states (SRT versus PS and SA), fact representation (FF versus SRT), and explicit derivations (CoT versus FF).

For each supervision target $c$, we train a separate LoRA adapter~\cite{hu2022lora} in the language layers of the VLM, keeping the vision encoder, connector, and base language weights frozen. Each adapted weight matrix $\mathbf W_0$ receives a trainable low-rank update $\Delta\mathbf W=\mathbf B\mathbf A$. Given an input $x=(I,q,\pi_c)$ with output format $\pi_c$, all six conditions minimize the same autoregressive loss over the training set $\mathcal D$,
\begin{equation}
\mathcal L_c(\omega) = -\mathbb E_{\mathcal D}\big[\log p_\omega(R_c\mid x)\big],
\label{eq:objective}
\end{equation}
where $p_\omega$ is the output distribution of the frozen VLM equipped with adapter $\omega$. For a target $R_c=\concat(\epsilon_c,y^\star)$ with prefix $\epsilon_c$, the loss decomposes into
\[
\mathcal L_c(\omega) = -\mathbb E_{\mathcal D}\big[
\underbrace{\log p_\omega(\epsilon_c\mid x)}_{\text{prefix generation}}
+ \underbrace{\log p_\omega(y^\star\mid x,\epsilon_c)}_{\text{conditioned answering}}\big].
\]
The second term conditions the answer on verified visual facts in SRT, FF, and CoT and on the placeholder in PS, whereas AO and SA answer without preceding states.

\subsection{Parallax: Learning When to Trust Reasoning}
\label{sec:parallax}

StateLens relies on ground-truth task states $z^\star$ for diagnosis and supervision, but natural VQA questions typically lack these annotations. Explicit reasoning can repair composition failures, yet it may also overturn correct direct answers or state wrong visual facts. As illustrated in Figure~\ref{fig:parallax}, \emph{Parallax} pairs the direct and reasoning routes, and their disagreement triggers source-bound verification against the image.

\noindent\textbf{Dual routes and disagreement gate.}
Given an image-question pair $(I,q)$, Parallax queries the same frozen VLM $M_\theta$ through two prompting routes. Direct predicts a concise answer $\hat y^{\rD}=M_\theta(I,q,\pi^{\rD})$, whereas Challenger produces $(\hat r^{\rC},\hat y^{\rC})=M_\theta(I,q,\pi^{\rC})$, with a draft $\hat r^{\rC}$ containing visual evidence and calculations before its final answer $\hat y^{\rC}$. Let $v^{\rD},v^{\rC}\in\{0,1\}$ indicate whether each answer is valid under the required output format. The disagreement gate is
\begin{equation}
G = v^{\rD}\,v^{\rC}\,\ind\big[\kappa(\hat y^{\rD})\neq\kappa(\hat y^{\rC})\big].
\label{eq:natural_gate}
\end{equation}
Parallax invokes the selector only when both routes return valid but distinct answers, and otherwise retains $\hat y^{\rD}$.

\noindent\textbf{Source-bound selector and order-averaged decision.}
When $G=1$, a LoRA selector $p_\eta$ on the same frozen VLM compares $\hat y^{\rD}$ and $\hat y^{\rC}$ as options $\mathrm A$ and $\mathrm B$ alongside $(I,q)$. Let $\tau$ map options $\{\mathrm A,\mathrm B\}$ to routes $\{\rD,\rC\}$ in one of the two orders, and let $b^\tau_s\in\{\mathrm A,\mathrm B\}$ be the option letter assigned to route $s$. The selector input is defined as
\begin{equation}
u_\tau = \big(I,\, q,\, \hat y^{\tau(\mathrm A)},\, \hat y^{\tau(\mathrm B)},\, h(b^\tau_{\rC})\big),
\label{eq:natural_input}
\end{equation}
where the source header $h(b^\tau_{\rC})$ states that the Challenger option comes with an unverified model-generated draft. The header moves with the Challenger answer across option orders, while the draft $\hat r^{\rC}$ is withheld so that the selector verifies both answers against the image.

\begin{table*}[!t]
\centering
\small
\setlength{\tabcolsep}{4pt}
\caption{Question-level diagnosis with StateLens. $\mathrm{E}{\to}\mathrm{S}$ feeds each model's Extract output directly into its Solve call without parameter updates. $|\Jset|$ counts jointly solvable questions, CF counts composition failures, and the bars partition all Direct errors.}
\label{tab:diagnosis}
\begin{tabular*}{\textwidth}{@{\extracolsep{\fill}}llrrrrrrrrc@{}}
\toprule
 & & \multicolumn{4}{c}{Accuracy (\%)} & \multicolumn{4}{c}{Composition failures} & Direct errors\\
\cmidrule(lr){3-6}\cmidrule(lr){7-10}\cmidrule(l){11-11}
Dataset & Model & Extract & Solve & Direct & $\mathrm{E}{\to}\mathrm{S}$ & $|\Jset|$ & CF & $\cfs$ (\%) & $\cfr$ (\%) & \raisebox{-1.6pt}{\includegraphics[width=1.62in]{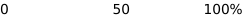}}\\
\midrule
\multirow{5}{*}{ChartQA} & Qwen2.5 & 81.1 & 86.8 & 63.5 & 75.1 & 770 & 158 & 40.2 & 20.5 & \raisebox{-1.6pt}{\includegraphics[width=1.62in]{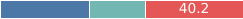}}\\
 & Qwen3.5 & 88.1 & 99.5 & 65.6 & 90.6 & 944 & 268 & 72.4 & 28.4 & \raisebox{-1.6pt}{\includegraphics[width=1.62in]{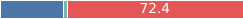}}\\
 & Molmo & 75.5 & 94.3 & 58.0 & 79.7 & 774 & 296 & 65.5 & 38.2 & \raisebox{-1.6pt}{\includegraphics[width=1.62in]{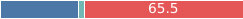}}\\
 & InternVL & 90.3 & 97.6 & 66.6 & 90.7 & 948 & 272 & 75.6 & 28.7 & \raisebox{-1.6pt}{\includegraphics[width=1.62in]{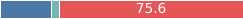}}\\
 & MiniCPM & 77.2 & 98.1 & 61.6 & 82.1 & 813 & 234 & 56.5 & 28.8 & \raisebox{-1.6pt}{\includegraphics[width=1.62in]{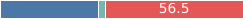}}\\
\midrule
\multirow{2}{*}{CLEVR} & Qwen2.5 & 56.3 & 94.5 & 37.5 & \textendash & 17{,}916 & 9{,}708 & 46.1 & 54.2 & \raisebox{-1.6pt}{\includegraphics[width=1.62in]{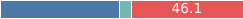}}\\
 & Molmo & 20.9 & 99.0 & 32.4 & \textendash & 7{,}014 & 4{,}044 & 17.7 & 57.7 & \raisebox{-1.6pt}{\includegraphics[width=1.62in]{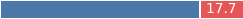}}\\
\bottomrule
\end{tabular*}
\par\vspace{2pt}
{\footnotesize
\textcolor[HTML]{4C78A8}{\rule{7pt}{5pt}}\ Extraction failure\hspace{10pt}
\textcolor[HTML]{72B7B2}{\rule{7pt}{5pt}}\ Solving failure\hspace{10pt}
\textcolor[HTML]{E45756}{\rule{7pt}{5pt}}\ Composition failure}
\end{table*}

On the training split, we collect gated questions where exactly one route is correct, $\mathcal T = \{i: G_i=1,\ c_i^{\rD}+c_i^{\rC}=1\}$, with $c_i^s\in\{0,1\}$ indicating route correctness and $s_i^\star\in\{\rD,\rC\}$ denoting the correct route. We present each training question in both orders $\tau$ and train the selector adapter $\eta$ by minimizing the cross-entropy loss
\begin{equation}
\mathcal L_{\mathrm{sel}}(\eta) = -\mathbb E_{i\in\mathcal T}\frac{1}{2}\sum_{\tau}\log p_\eta\big(b^\tau_{s_i^\star},\mathrm{EOS}\mid u_{i,\tau}\big),
\label{eq:natural_loss}
\end{equation}
where $\mathrm{EOS}$ is the end-of-sequence token. At test time, we normalize the next-token probabilities over $\mathrm A$ and $\mathrm B$ into $P_\eta(b\mid u_\tau)$, average the Challenger probability across both orders to reduce position bias~\cite{zheng2023judging}, and select $\hat y$ via
\begin{equation}
\bar p^{\rC} = \frac{1}{2}\sum_{\tau} P_\eta\big(b^\tau_{\rC}\mid u_\tau\big),\;\;
\hat y = \begin{cases}
\hat y^{\rC}, & \bar p^{\rC}\ge 0.5,\\
\hat y^{\rD}, & \text{otherwise.}
\end{cases}
\label{eq:natural_switch}
\end{equation}
Parallax thus keeps the Direct answer unless the selector favors Challenger on average over both orders.

\begin{table}[!t]
\centering
\footnotesize
\setlength{\tabcolsep}{2.2pt}
\caption{Diagnosis across Mixture task structures.}
\label{tab:mix_diag}
\begin{tabular*}{\columnwidth}{@{\extracolsep{\fill}}lrrrrrrrrr@{}}
\toprule
 & \multicolumn{3}{c}{Qwen2.5} & \multicolumn{3}{c}{Qwen3.5} & \multicolumn{3}{c}{Molmo}\\
\cmidrule(lr){2-4}\cmidrule(lr){5-7}\cmidrule(l){8-10}
Structure & $|\Jset|$ & CF & $\cfr$ & $|\Jset|$ & CF & $\cfr$ & $|\Jset|$ & CF & $\cfr$\\
\midrule
Arithmetic & 26 & 25 & 96.2 & 100 & 0 & 0.0 & 57 & 49 & 86.0\\
Lookup & 65 & 33 & 50.8 & 118 & 0 & 0.0 & 4 & 2 & 50.0\\
Ranking & 35 & 28 & 80.0 & 62 & 47 & 75.8 & 0 & 0 & \textendash\\
Color set & 151 & 68 & 45.0 & 221 & 25 & 11.3 & 44 & 25 & 56.8\\
\midrule
Overall & 277 & 154 & 55.6 & 501 & 72 & 14.4 & 105 & 76 & 72.4\\
\bottomrule
\end{tabular*}
\end{table}

\begin{table}[!t]
\centering
\footnotesize
\setlength{\tabcolsep}{2.8pt}
\caption{Comparison of supervision targets across benchmarks.}
\label{tab:targets}
\begin{tabular*}{\columnwidth}{@{\extracolsep{\fill}}llrrrrrr@{}}
\toprule
Dataset & Model & AO & PS & SA & SRT & FF & CoT\\
\midrule
\multirow{5}{*}{ChartQA} & Qwen2.5 & 86.07 & 85.24 & 84.87 & 92.01 & \textbf{92.66} & 92.48\\
 & Qwen3.5 & 88.02 & 88.02 & 88.49 & \textbf{94.99} & 94.43 & 94.24\\
 & Molmo & 77.90 & 75.86 & 80.41 & \textbf{91.83} & 91.18 & 90.81\\
 & InternVL & 86.72 & 86.17 & 87.65 & 93.69 & 93.78 & \textbf{93.87}\\
 & MiniCPM & 73.07 & 71.12 & 73.82 & 85.70 & \textbf{85.79} & 85.42\\
\midrule
\multirow{3}{*}{Mixture} & Qwen2.5 & 70.21 & 70.14 & 67.46 & 88.82 & 90.05 & \textbf{93.26}\\
 & Qwen3.5 & 79.02 & 79.10 & 80.32 & 93.95 & \textbf{94.49} & \textbf{94.49}\\
 & Molmo & 50.46 & 48.55 & 57.43 & 81.55 & 83.61 & \textbf{86.29}\\
\midrule
\multirow{2}{*}{CLEVR} & Qwen2.5 & 93.70 & 92.91 & 93.53 & \textbf{95.39} & \textendash & \textendash\\
 & Molmo & 87.74 & 86.80 & 90.64 & \textbf{92.02} & \textendash & \textendash\\
\bottomrule
\end{tabular*}
\end{table}

\section{Experiments}
\label{sec:experiments}

\subsection{Datasets and Implementation Details}
\label{sec:setup}

\noindent\textbf{Datasets.}
Following Eq.~\eqref{eq:construction}, StateLens constructs three benchmarks with doubly verified task states. The constructed \emph{ChartQA}~\cite{masry2022chartqa} contains $9{,}993$ training and $1{,}077$ test questions from $3{,}331$ and $359$ charts, respectively, under the official split. Each question applies comparison, difference, or sum to a task state of two verified table-cell values. \emph{Mixture} contains $4{,}000$ training and $1{,}306$ test questions from four task families, namely multi-step arithmetic over three chart values ($291$ test), conditional lookup in a $3\times2$ chart table ($215$), relational count ranking over three spatial groups ($400$), and intersection or difference of two color sets ($400$). The constructed \emph{CLEVR}~\cite{johnson2017clevr} contains $158{,}226$ training and $33{,}723$ test questions, applying the same three operations as in StateLens ChartQA to two verified count queries. For Parallax, we use the official \emph{ChartQA} benchmark~\cite{masry2022chartqa} with $28{,}299$ training and $2{,}500$ test questions, where half of the test questions are written by humans.

\noindent\textbf{Models and evaluation protocols.}
We evaluate Qwen2.5-VL-7B~\cite{bai2025qwen25vl}, Qwen3.5-9B~\cite{qwenteam2026qwen35}, Molmo2-O-7B~\cite{clark2026molmo2}, InternVL3.5-8B~\cite{wang2025internvl35}, and MiniCPM-V~4.5~\cite{yu2025minicpmv45}, denoted as Qwen2.5, Qwen3.5, Molmo, InternVL, and MiniCPM. StateLens tasks use exact canonical matching alongside $\cfs$ and $\cfr$ in Eq.~\eqref{eq:metrics}, whereas original ChartQA uses the official relaxed accuracy~\cite{masry2022chartqa} within $5\%$ for numeric answers. Relative to baseline errors $\mathcal R^{(0)}$ ($\Fcmp^{(0)}$ in StateLens and Direct errors in Parallax) and baseline Direct successes $\Gset^{(0)}$, we track repairs $N_{\mathrm{rep}} = \sum_{i\in\mathcal R^{(0)}} \hat c_i$ and regressions $N_{\mathrm{reg}} = \sum_{i\in\Gset^{(0)}} (1 - \hat c_i)$ for correctness $\hat c_i\in\{0,1\}$.

\noindent\textbf{Implementation details.}
All LoRA adapters use rank $16$, scaling factor $32$, and dropout $0.05$ in the language layers. We train them for one epoch with AdamW~\cite{loshchilov2019decoupled} (learning rate $10^{-4}$, batch size $8$, seed $42$) and use deterministic decoding at inference. For each model, supervision variants share the same training instances, data order, and optimization settings, differing only in output instructions and targets. For Parallax on original ChartQA, we use frozen Qwen3.5 with thinking disabled and train the selector on $4{,}433$ eligible questions. The Molmo selector is trained in the same way on $12{,}272$ eligible questions with its own candidates. Additional random seeds, prompt templates, and scoring protocols are provided in the supplementary material.

\begin{figure*}[!t]
\centering
\includegraphics[width=0.92\textwidth]{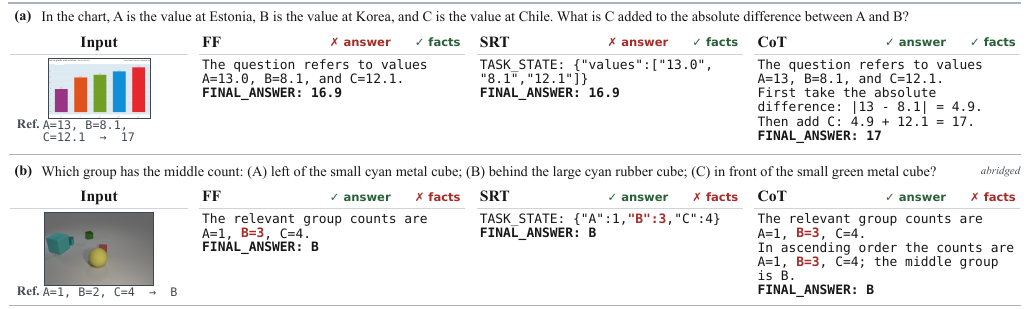}
\caption{Qualitative examples of written facts and final answers on Mixture with Qwen2.5. (a) On multi-step arithmetic, FF and SRT read all three values accurately yet miss the two-step calculation, whereas CoT writes the intermediate step and answers correctly. (b) On relational count ranking, all three targets miscount group B yet still output the correct middle group.}
\label{fig:qualitative}
\end{figure*}

\subsection{Diagnosis of Composition Failures}
\label{sec:exp_diagnosis}

\noindent\textbf{Prevalence across models and domains.}
We first evaluate the untuned VLMs on StateLens ChartQA and CLEVR via Extract, Solve, and Direct. As shown in Table~\ref{tab:diagnosis}, Direct accuracy on StateLens ChartQA ranges from 58.0\% to 66.6\% across five VLMs, compared with 75.5--90.3\% for Extract and 86.8--99.5\% for Solve. For Qwen2.5, the Fr\'echet bounds only place $\rho$ between 4.4\% and 36.5\%, whereas question-level diagnosis identifies 14.7\%. Across the five VLMs, composition failures account for 40.2--75.6\% of Direct errors, and Direct fails on 20.5--38.2\% of jointly solvable questions. On CLEVR, this rate rises to 54.2\% and 57.7\% for Qwen2.5 and Molmo, respectively. Feeding each model's Extract output into its Solve call ($\mathrm{E}{\to}\mathrm{S}$) improves ChartQA accuracy over Direct by 11.6--25.0 points without parameter updates.

\noindent\textbf{Richer task structures and natural questions.}
Table~\ref{tab:mix_diag} shows that composition failures persist on Mixture, with overall $\cfr$ values of 55.6\%, 14.4\%, and 72.4\% for Qwen2.5, Qwen3.5, and Molmo, respectively. On arithmetic and lookup, Qwen3.5 writes brief calculations under Direct and has a $\cfr$ of 0.0\%, yet it still fails on 75.8\% of jointly solvable ranking questions. We further diagnose 443 human-written questions from the original ChartQA test split, whose task-state annotation is described in the supplementary material. Composition failures still occur on 20 questions for Qwen2.5 and 11 for Molmo, accounting for 23.0\% and 13.3\% of their Direct errors.

\begin{table}[!t]
\centering
\footnotesize
\setlength{\tabcolsep}{1.4pt}
\caption{Comparison across Mixture task structures.}
\label{tab:family}
\begin{tabular*}{\columnwidth}{@{\extracolsep{\fill}}lrrrrrrrrrrrr@{}}
\toprule
 & \multicolumn{4}{c}{Qwen2.5} & \multicolumn{4}{c}{Qwen3.5} & \multicolumn{4}{c}{Molmo}\\
\cmidrule(lr){2-5}\cmidrule(lr){6-9}\cmidrule(l){10-13}
Structure & AO & SRT & FF & CoT & AO & SRT & FF & CoT & AO & SRT & FF & CoT\\
\midrule
Arith. & 32.3 & 67.7 & 68.4 & \textbf{83.5} & 50.9 & 84.5 & 87.3 & \textbf{89.0} & 22.7 & 68.7 & 68.7 & \textbf{82.8}\\
Lookup & 71.6 & 85.1 & 87.4 & \textbf{88.8} & 80.0 & 91.6 & \textbf{92.6} & \textbf{92.6} & 67.4 & 84.2 & 84.7 & \textbf{86.5}\\
Ranking & 87.8 & 96.5 & \textbf{97.8} & 97.3 & 91.0 & \textbf{97.5} & 95.5 & 97.0 & 38.5 & 80.0 & \textbf{81.5} & 80.0\\
Color & 79.5 & 98.5 & \textbf{99.5} & 98.8 & 87.0 & 98.5 & \textbf{99.8} & 97.0 & 73.5 & 91.0 & \textbf{96.0} & 95.0\\
\bottomrule
\end{tabular*}
\end{table}

\begin{table}[!t]
\centering
\footnotesize
\setlength{\tabcolsep}{1.1pt}
\caption{Outcomes beyond final accuracy. Repair (\%) on $\Fcmp^{(0)}$, Regressions on $\Gset^{(0)}$, and correct answers with wrong facts (\%).}
\label{tab:beyond}
\begin{tabular*}{\columnwidth}{@{\extracolsep{\fill}}llrrrrrrrrrrr@{}}
\toprule
 & & \multicolumn{4}{c}{Repair (\%) $\uparrow$} & \multicolumn{4}{c}{Regressions $\downarrow$} & \multicolumn{3}{c}{Wrong fact (\%) $\downarrow$}\\
\cmidrule(lr){3-6}\cmidrule(lr){7-10}\cmidrule(l){11-13}
Data & Model & AO & SRT & FF & CoT & AO & SRT & FF & CoT & SRT & FF & CoT\\
\midrule
\multirow{5}{*}{Chart} & Qwen2.5 & 77.8 & 94.9 & \textbf{97.5} & 96.8 & 26 & 8 & \textbf{6} & 8 & 3.6 & 3.3 & \textbf{3.1}\\
 & Qwen3.5 & 82.5 & \textbf{99.6} & 97.8 & 98.5 & 19 & 3 & \textbf{2} & 4 & \textbf{2.7} & \textbf{2.7} & \textbf{2.7}\\
 & Molmo & 71.3 & 97.0 & \textbf{98.6} & 97.0 & 48 & \textbf{5} & 13 & 7 & \textbf{3.7} & 3.8 & 3.9\\
 & InternVL & 80.9 & \textbf{97.8} & 97.4 & \textbf{97.8} & 20 & 6 & \textbf{3} & \textbf{3} & \textbf{2.8} & \textbf{2.8} & 2.9\\
 & MiniCPM & 58.5 & 90.2 & 92.3 & \textbf{92.7} & 43 & 20 & \textbf{17} & 20 & \textbf{6.0} & \textbf{6.0} & 6.3\\
\midrule
\multirow{3}{*}{Mix.} & Qwen2.5 & 74.0 & 93.5 & 94.2 & \textbf{96.8} & 60 & 13 & 10 & \textbf{5} & 8.4 & \textbf{8.2} & 10.5\\
 & Qwen3.5 & 88.9 & 97.2 & \textbf{98.6} & 94.4 & 167 & 20 & 10 & \textbf{7} & \textbf{6.9} & 10.8 & 9.6\\
 & Molmo & 61.8 & \textbf{97.4} & 94.7 & \textbf{97.4} & 40 & 10 & 9 & \textbf{8} & 18.2 & \textbf{15.7} & 17.3\\
\bottomrule
\end{tabular*}
\end{table}

\subsection{Decoupling Supervision with StateLens}
\label{sec:exp_dissect}

\noindent\textbf{Content and position of the task state.}
Table~\ref{tab:targets} compares AO, PS, SA, and SRT across ChartQA, Mixture, and CLEVR. PS gains at most 0.08 points over AO across the ten model--benchmark settings, and SA stays within 2.9 points of AO in nine of them. In contrast, SRT outperforms the better of PS and SA by 6.0--11.9 points on ChartQA, 13.6--24.1 on Mixture, and 1.4--1.9 on CLEVR. It also surpasses the two-call $\mathrm{E}{\to}\mathrm{S}$ baseline by 3.0--16.9 points in a single response. The gains thus come from verified pre-answer facts rather than output formatting or post-answer supervision.

\noindent\textbf{Visual facts, form, and derivations.}
As shown in Tables~\ref{tab:targets} and~\ref{tab:family}, placing visual facts before the answer improves accuracy over AO by 5.9--13.9 points across the five ChartQA models and by 14.9--35.8 points on Mixture. On ChartQA, FF, SRT, and CoT differ by at most 1.1 points within each model. Across ChartQA and Mixture, FF and SRT differ by at most 2.1 points, while SRT uses roughly half as many output tokens as CoT (17.8--29.0 versus 42.6--58.5). The benefit of explicit derivations is task-dependent. On Mixture, CoT differs from FF by $-2.8$ to $+1.9$ points on lookup, ranking, and color-set tasks. In contrast, CoT achieves the highest accuracy on multi-step arithmetic for all three models, outperforming FF by 15.1 points on Qwen2.5 and 14.1 on Molmo. Figure~\ref{fig:qualitative}(a) shows a typical case.

\begin{table}[!t]
\centering
\footnotesize
\setlength{\tabcolsep}{2.4pt}
\caption{ChartQA comparison. CoT prompt adapted from~\cite{masry2025chartqapro}.}
\label{tab:natural_comparison}
\begin{tabular*}{\columnwidth}{@{\extracolsep{\fill}}llcr@{}}
\toprule
Type & Method & Venue & Acc. (\%)\\
\midrule
\multirow{12}{*}{Chart-specific} & MatCha\tablecite{liu2023matcha} & ACL'23 & 64.2\\
 & UniChart\tablecite{masry2023unichart} & EMNLP'23 & 66.2\\
 & ChartAst-S\tablecite{meng2024chartassistant} & Find. ACL'24 & 79.9\\
 & TinyChart@768\tablecite{zhang2024tinychart} & EMNLP'24 & 83.6\\
 & EvoChart\tablecite{huang2025evochart} & AAAI'25 & 81.5\\
 & ChartMoE+PoT\tablecite{chartmoe2025} & ICLR'25 & 84.6\\
 & ECD\tablecite{yang2025ecd} & ICCV'25 & 85.3\\
 & ChartPoint\tablecite{xu2025chartpoint} & ICCV'25 & 87.7\\
 & BigCharts-R1\tablecite{bigchartsr1_2025} & COLM'25 & 89.8\\
 & VisDoT (short)\tablecite{lee2026visdot} & Find. EACL'26 & 85.6\\
 & START-RL\tablecite{liu2026start} & WACV'26 & 88.8\\
 & Chart-FR1\tablecite{pan2026chartfr1} & CVPR'26 & \underline{91.0}\\
\midrule
\multirow{6}{*}{General VLM} & GPT-4V\tablecite{openai2023gpt4,zhang2024tinychart} & OpenAI'23 & 78.5\\
 & GPT-4o\tablecite{openai2024gpt4o,anthropic2024claude35card} & OpenAI'24 & 85.7\\
 & Gemini-1.5-Pro\tablecite{gemini2024gemini15} & arXiv'24 & 87.2\\
 & InternVL2.5-78B\tablecite{chen2024internvl25} & arXiv'24 & 88.3\\
 & Qwen2.5-VL-72B\tablecite{bai2025qwen25vl} & arXiv'25 & 89.5\\
 & Claude-3.5-Sonnet\tablecite{anthropic2024claude35,anthropic2024claude35card} & Anthropic'24 & 90.8\\
\midrule
\multirow{4}{*}{Qwen3.5} & Direct & \textendash & 86.68\\
 & Step-by-step CoT & \textendash & 78.64\\
 & Thinking mode & \textendash & 75.12\\
 & \textbf{Parallax (ours)} & \textendash & \textbf{91.44}\\
\bottomrule
\end{tabular*}
\end{table}

\noindent\textbf{Failure repair and reliability of written facts.}
With $\Fcmp^{(0)}$ and $\Gset^{(0)}$ fixed before tuning, Table~\ref{tab:beyond} shows that SRT, FF, and CoT repair 90.2--99.6\% of composition failures and reduce regressions from 19--167 under AO to at most 20. On Mixture, most additional correct predictions of CoT over SRT come from baseline Direct errors outside $\Fcmp^{(0)}$, such as 45 of 58 on Qwen2.5. However, correct answers do not guarantee accurate visual facts. Among correct predictions, 2.7--6.3\% on ChartQA and 6.9--18.2\% on Mixture contain incorrect facts, and CoT does not consistently reduce this rate relative to FF. Figure~\ref{fig:qualitative}(b) shows such a case.

\subsection{Evaluation of Parallax on ChartQA}
\label{sec:exp_parallax}

\noindent\textbf{Comparison with state-of-the-art methods.}
We evaluate Parallax on all $2{,}500$ original ChartQA test questions. As shown in Table~\ref{tab:natural_comparison}, Parallax achieves 91.44\% accuracy on Qwen3.5, outperforming Direct, Step-by-step CoT, and Thinking mode by 4.76, 12.80, and 16.32 points, respectively. Under each work's reported protocol, it also exceeds Chart-FR1 (91.0\%) and Claude-3.5-Sonnet (90.8\%), the strongest chart-specific model and general VLM in the table. Moreover, Step-by-step CoT and Thinking mode generate about $3\times$ and $13\times$ as many tokens as Parallax, which averages $76.6$ tokens per question. Most of them come from Challenger, and the selector scores only the $693$ gated questions ($27.7\%$) with two extra next-token passes.

\noindent\textbf{Ablation study.}
Table~\ref{tab:natural_ablation} compares the routes, selector designs, and test-time inputs. Applying Challenger to every question repairs 131 Direct errors at the cost of 513 regressions, and 129 of these repairs have valid answers from both routes. Without training, both selector scoring and a zero-shot judge that reads the draft~\cite{zheng2023judging} incur at least 50 regressions. A trained symmetric selector reduces regressions to 5 but captures only 100 repairs. With the source header and the draft withheld, Parallax captures 125 of the 129 valid repairs with 6 regressions, while exposing the full draft brings no gain. Reversing the header cuts the repairs of Parallax to 16, and a mismatched draft cuts those of the full-draft variant from 122 to 53. Removing or mismatching the image and hiding the question keep the repairs but raise regressions to 34--48, suggesting that the selector mainly uses both inputs to protect correct Direct answers.

\begin{table}[!t]
\centering
\footnotesize
\setlength{\tabcolsep}{2.4pt}
\caption{Ablation study of Parallax on ChartQA.}
\label{tab:natural_ablation}
\begin{tabular*}{\columnwidth}{@{\extracolsep{\fill}}llrrr@{}}
\toprule
Group & Setting & Acc. (\%) & Repairs $\uparrow$ & Regr. $\downarrow$\\
\midrule
\multirow{2}{*}{Routes} & Direct route (anchor) & 86.68 & \textendash & \textendash\\
 & Challenger route & 71.40 & 131 & 513\\
\midrule
\multirow{5}{*}{Selector} & Untrained scoring & 86.16 & 37 & 50\\
 & Zero-shot judge\tablecite{zheng2023judging} & 88.44 & 108 & 64\\
 & Symmetric (w/o header) & 90.48 & 100 & 5\\
 & Source-bound + full draft & 91.32 & 122 & 6\\
 & \textbf{Parallax (ours)} & \textbf{91.44} & \textbf{125} & \textbf{6}\\
\midrule
\multirow{5}{*}{Probes} & w/o image & 90.28 & 127 & 37\\
 & w/ mismatched image & 89.84 & 127 & 48\\
 & w/o question & 90.32 & 125 & 34\\
 & w/ reversed header & 87.32 & 16 & 0\\
 & w/ mismatched draft & 88.72 & 53 & 2\\
\bottomrule
\end{tabular*}
\end{table}

\begin{table}[!t]
\centering
\footnotesize
\setlength{\tabcolsep}{4pt}
\caption{Generalization across question distributions and VLMs.}
\label{tab:bridge}
\begin{tabular*}{\columnwidth}{@{\extracolsep{\fill}}llr@{}}
\toprule
Setting & Method & Acc. (\%)\\
\midrule
\multirow{4}{*}{\begin{tabular}[c]{@{}l@{}}StateLens\\ChartQA\\(Qwen3.5)\end{tabular}} & Direct & 65.65\\
 & Zero-shot state-first prompt & 66.67\\
 & Two-call $\mathrm{E}{\to}\mathrm{S}$ & 90.62\\
 & \textbf{Parallax (ours)} & \textbf{91.27}\\
\midrule
\multirow{4}{*}{\begin{tabular}[c]{@{}l@{}}ChartQA\\(Molmo)\end{tabular}} & Direct & 58.00\\
 & Step-by-step CoT & 64.16\\
 & Parallax (Molmo selector) & 79.20\\
 & Parallax (Qwen3.5 selector) & \textbf{79.96}\\
\bottomrule
\end{tabular*}
\end{table}

\noindent\textbf{Generalization across question distributions and VLMs.}
We first apply the selector trained on original ChartQA to StateLens ChartQA without retraining. As shown in Table~\ref{tab:bridge}, Parallax raises Qwen3.5 accuracy from 65.65\% to 91.27\% without task-state supervision, surpassing the two-call $\mathrm{E}{\to}\mathrm{S}$ baseline. Among the 268 composition failures in Table~\ref{tab:diagnosis}, it recovers 256 of the 259 repairable cases. We further apply Parallax to Molmo on original ChartQA, where valid disagreements rise from 693 to $1{,}662$ questions. A selector trained on Molmo candidates repairs 538 Direct errors with 8 regressions, improving accuracy by 18.20 points over the untrained selector. The frozen Qwen3.5 selector captures 560 of the 563 valid repairs with 11 regressions. Both raise Molmo accuracy by more than 21 points over Direct.

\section{Conclusion}
\label{sec:conclusion}

In this paper, we investigate why VLMs that extract visual facts and solve from them separately still fail when answering directly. We propose StateLens, which uses doubly verified task states to diagnose composition failures per question and decouple intermediate supervision. Across five VLMs, composition failures account for a major share of Direct errors and also occur on human-written questions. Pre-answer visual facts repair most of these failures, explicit derivations mainly aid multi-step arithmetic, and correct answers do not certify the written facts. Motivated by these findings, we propose Parallax to verify candidate answers against the image with source awareness, which improves direct answering on both Qwen3.5 and Molmo and transfers to StateLens ChartQA without retraining. In the future, we hope to extend source-bound verification to open-ended multimodal reasoning beyond charts and scenes.

\section*{LLM Usage Statement}

OpenAI ChatGPT, OpenAI Codex, and Anthropic Claude (via Cursor) assisted with editing and reorganizing text, checking literature and citations, and refining question templates, verification rules, result analyses, and figure layouts. ChatGPT also provided reviewer-style feedback. The authors reviewed and edited AI-generated scripts for data construction, evaluation, analysis, typesetting, and plotting. OpenAI image generation supplied icon assets adapted in Figure~\ref{fig:parallax}. The authors take full responsibility for the submitted content, including all AI-assisted parts.

{\small
\fontsize{9}{10.1}\selectfont
\setlength{\bibsep}{0.2pt plus 0.2pt}
\bibliographystyle{ieeenat_fullname}
\bibliography{main}
}

\clearpage
\onecolumn
\raggedbottom
\setcounter{topnumber}{4}
\setcounter{bottomnumber}{3}
\setcounter{totalnumber}{6}
\renewcommand{\topfraction}{0.92}
\renewcommand{\bottomfraction}{0.78}
\renewcommand{\textfraction}{0.06}
\renewcommand{\floatpagefraction}{0.75}
\setlength{\textfloatsep}{12pt plus 3pt minus 2pt}
\setlength{\floatsep}{10pt plus 3pt minus 2pt}
\setlength{\intextsep}{10pt plus 3pt minus 2pt}
\begin{center}
{\Large\bfseries\thetitle\par}
\vspace{0.5em}
{\large Supplementary Material\par}
\end{center}
\setcounter{section}{0}
\setcounter{subsection}{0}
\setcounter{subsubsection}{0}
\setcounter{table}{0}
\setcounter{figure}{0}
\setcounter{equation}{0}
\setcounter{algorithm}{0}
\providecommand{\theHsection}{}
\renewcommand{\theHsection}{supp.\arabic{section}}
\providecommand{\theHsubsection}{}
\renewcommand{\theHsubsection}{supp.\arabic{section}.\arabic{subsection}}
\providecommand{\theHsubsubsection}{}
\renewcommand{\theHsubsubsection}{supp.\arabic{section}.\arabic{subsection}.\arabic{subsubsection}}
\providecommand{\theHtable}{}
\renewcommand{\theHtable}{supp.\arabic{table}}
\providecommand{\theHfigure}{}
\renewcommand{\theHfigure}{supp.\arabic{figure}}
\providecommand{\theHequation}{}
\renewcommand{\theHequation}{supp.\arabic{equation}}
\providecommand{\theHalgorithm}{}
\renewcommand{\theHalgorithm}{supp.\arabic{algorithm}}
\renewcommand{\thesection}{S\arabic{section}}
\renewcommand{\thetable}{S\arabic{table}}
\renewcommand{\thefigure}{S\arabic{figure}}
\renewcommand{\theequation}{S\arabic{equation}}
\renewcommand{\thealgorithm}{S\arabic{algorithm}}

This supplementary material provides formal proofs, benchmark construction procedures, prompt templates, and extended analyses supporting the main paper. Section~\ref{sec:s-construction} derives the Fr\'echet bounds and details the reverse construction of StateLens benchmarks. Section~\ref{sec:s-protocol} documents the prompt templates, output formats, and scoring rules for both StateLens and Parallax. Section~\ref{sec:s-diagnosis} extends the question-level composition-failure diagnosis and robustness checks. Section~\ref{sec:s-training} reports multi-seed, paired-bootstrap, and written-fact analyses across supervision targets. Section~\ref{sec:s-arbitration} details Parallax training, ablations, and transfer evaluations. Finally, Section~\ref{sec:s-examples} presents additional qualitative examples.

\section{Formal Proof and Reverse Construction}
\label{sec:s-construction}

\subsection{Proof of the Fr\'echet Bounds}
Let $A=\{i:E_i=1\}$, $B=\{i:S_i=1\}$, and $C=\{i:D_i=0\}$ over $\{1,\dots,N\}$, with cardinalities $|A|=N\bar E$, $|B|=N\bar S$, and $|C|=N(1-\bar D)$. By definition, $N\rho=|A\cap B\cap C|$. Because $A\cap B\cap C$ is a subset of each of $A$, $B$, and $C$, we have $N\rho\le\min(|A|,|B|,|C|)=N\min(\bar E,\bar S,1-\bar D)$. For the lower bound, De Morgan's law and subadditivity give
\begin{equation}
\begin{aligned}
|A\cap B\cap C|&=N-|A^c\cup B^c\cup C^c|\\
&\ge N-\min(N,3N-|A|-|B|-|C|)\\
&=\max(0,|A|+|B|+|C|-2N),
\end{aligned}
\end{equation}
and dividing by $N$ yields $\rho\ge\max(0,\bar E+\bar S-\bar D-1)$. Both bounds are attainable for any integer counts $(|A|,|B|,|C|)\in\{0,\dots,N\}^3$. To attain the upper bound, start each of $A$, $B$, and $C$ at index 1 so that the smallest set is contained in the other two. To attain the lower bound, arrange the $N$ indices on a circle and assign consecutive blocks of lengths $|A^c|$, $|B^c|$, and $|C^c|$, wrapping at $N$ if necessary. The blocks cover $\min(N,3N-|A|-|B|-|C|)$ distinct indices without changing any set size.

\subsection{ChartQA and CLEVR Pair-State Construction}
Both constructors operate solely on the structured dataset annotations $\mathcal A$ prior to any model evaluation. From the verified subset $\mathcal A^\star$, labels or count queries form $q$, values or counts form $z^\star$, and the reference answer is computed deterministically as $y^\star=\phi_o(z^\star)$.

On ChartQA~\cite{masry2022chartqa}, each chart supplies an image, a source table, and a structured chart annotation covering vertical-bar, horizontal-bar, line, and pie charts. Labels undergo Unicode NFKC normalization, case folding, whitespace collapsing, and canonical quotation and dash mapping. Numeric cells are parsed as finite decimals after stripping commas and any trailing percent symbol or word \texttt{percent}, without rescaling. Duplicate annotation points are merged, and a generic single-series annotation label such as \texttt{bar} is matched to a single-series table. A table cell is verified only when a unique annotation point matches its normalized series and position and the two values satisfy
\begin{equation}
|v_{\mathrm{table}}-v_{\mathrm{annotation}}|\le\max(10^{-6},10^{-4}|v_{\mathrm{table}}|).
\label{eq:s-tolerance}
\end{equation}
This tolerance is used only when cross-checking the two dataset annotation sources, whereas all model predictions are canonicalized and compared exactly. Algorithm~\ref{alg:s-chartqa} selects the table orientation with the strictly largest number of verified cells, forms candidate pairs sorted by normalized $(\text{series},\text{position})$ keys, and deterministically picks one pair per chart using the smallest SHA256 digest of the chart identifier joined with the two cell keys. Any chart involved in a cross-split image or table SHA256 collision is removed from all affected splits. Official train and test splits are preserved without merging validation data into training.

\begin{algorithm}[!htbp]
\caption{ChartQA question and state construction}
\label{alg:s-chartqa}
\small
\begin{algorithmic}[1]
\Require Official ChartQA train, validation, and test splits
\For{each chart in each split}
 \State Require a readable image, rectangular table, and supported chart annotation
 \State Parse both table orientations and normalize labels and finite decimal values
 \State Parse annotation series, positions, and values and merge exact duplicates
 \For{each orientation with unique normalized cell keys}
  \State Match table cell labels to annotation series and positions
  \State Keep cells with a unique match satisfying Eq.~\eqref{eq:s-tolerance}
 \EndFor
 \State Choose the unique orientation with the largest verified-cell count
 \State Reject tied orientations and charts with fewer than two verified cells
 \State Sort verified cells by normalized $(\text{series},\text{position})$ key
 \State Select the unordered pair with the smallest SHA256 pair key in lexicographic cell order
 \State Set $z^\star=(a,b)$ to the two verified values
 \For{$o$ in comparison, absolute difference, sum}
  \State Construct $q=\psi_q(\ell(\mathcal A^\star),o)$ and compute $y^\star=\phi_o(z^\star)$
  \State Store $(I,q,z^\star,y^\star)$ with source identifiers and verification record
 \EndFor
\EndFor
\State Remove all charts with image or table SHA256 collisions across official splits
\end{algorithmic}
\end{algorithm}

On CLEVR~\cite{johnson2017clevr}, we use the official train split for training and the official validation split as the test set. Every official question whose functional program ends with \texttt{count} is re-executed on the scene graph by a symbolic executor supporting attribute filters, unique-object selection, spatial relations, same-attribute queries, unions, intersections, and counts. Any program whose re-executed count disagrees with the official answer is discarded. As summarized in Algorithm~\ref{alg:s-clevr}, among scenes with at least two verified count questions having distinct program hashes, one ordered pair is chosen deterministically via SHA256 hashing of the split, image index, and question indices. Each constructed question states \texttt{First count query:} and \texttt{Second count query:} verbatim, followed by comparison, absolute difference, or sum over the two counts.

\begin{algorithm}[!htbp]
\caption{CLEVR question and state construction}
\label{alg:s-clevr}
\small
\begin{algorithmic}[1]
\Require Official CLEVR images, scene graphs, questions, answers, and programs
\For{each official question in train or validation}
 \If{the final program operation is \texttt{count}}
  \State Re-execute the full program on the scene graph
  \State Retain the question if the executed count equals the official answer
 \EndIf
\EndFor
\For{each scene with at least two retained questions having distinct program hashes}
 \State Sort retained questions by official question index
 \State Select the distinct-program pair with the smallest SHA256 identity key
 \State Set $z^\star=(a,b)$ to the two verified counts in question-index order
 \For{$o$ in comparison, absolute difference, sum}
  \State Combine both count queries with operation $o$ to form $q$ and compute $y^\star=\phi_o(z^\star)$
  \State Store $(I,q,z^\star,y^\star)$ with program hashes and verification record
 \EndFor
\EndFor
\end{algorithmic}
\end{algorithm}

\subsection{Construction of the Four Mixture Families}
To test richer state structures and multi-step operations, four additional families are constructed from verified ChartQA tables and CLEVR scene graphs:
\begin{itemize}[leftmargin=*,topsep=2pt,itemsep=2pt]
\item \textbf{Multi-step arithmetic (ChartQA, $1{,}000$ train / $291$ test).} Three verified cells $(a,b,c)$ are selected per chart and paired with composed arithmetic operations such as $(a+b)-c$ and $|a-b|+c$. The state $z^\star$ stores the three chart values in question order without intermediate results.
\item \textbf{Conditional lookup (ChartQA, $1{,}000$ train / $215$ test).} Three series $A,B,C$ are evaluated at two positions $P,Q$, forming a $3\times2$ table of six verified values with a unique maximum at $P$. The question asks for the value at $Q$ of the series that is largest at $P$, and $z^\star$ stores all six entries.
\item \textbf{Relational count ranking (CLEVR, $1{,}000$ train / $400$ test).} Three object groups $A,B,C$ are defined on the scene graph through spatial relations to uniquely identifiable reference objects, with three distinct counts. The question asks which group has the middle count (neither largest nor smallest), and $z^\star$ stores the three counts in $(A,B,C)$ order.
\item \textbf{Color-set operations (CLEVR, $1{,}000$ train / $400$ test).} Two object groups are defined on the scene graph, and $z^\star$ records the sorted list of distinct colors present in each group. The question asks for either the color intersection or the set difference between the two groups, reported as a comma-separated list or \texttt{none}.
\end{itemize}
The ChartQA families use official train and test splits with the cross-split image and table hash deduplication of Section~S1.2, while the CLEVR families use official train and validation scenes for training and testing, respectively. For ChartQA and all Mixture families, we verbalize the verified entries of $z^\star$ as $e^\star$ without computing the answer, then execute $\phi_o$ to obtain $d^\star$. The derivation records intermediate arithmetic results, selects the series at $P$ before reading it at $Q$, orders counts, or applies the requested set operation. FF and CoT share $e^\star$; only CoT adds $d^\star$.

\FloatBarrier
\section{Prompts, Output Formats, and Scoring Rules}
\label{sec:s-protocol}

\subsection{Inputs and Outputs Across Interfaces and Targets}
As summarized in Table~\ref{tab:s-conditions}, every condition except Solve receives the image and the full question, and only Solve receives the ground-truth state $z^\star$ as part of its input. During fine-tuning, the output column specifies the supervised assistant response $R_c$, whereas at evaluation time the model generates the entire response itself.

\begin{table}[!htbp]
\centering
\small
\setlength{\tabcolsep}{3pt}
\caption{\textbf{Inputs and outputs of all diagnostic interfaces and supervision targets.} State supplied indicates whether $z^\star$ is provided in the prompt. At test time, every fine-tuned adapter generates its entire output without ground-truth state input.}
\label{tab:s-conditions}
\begin{tabular*}{\textwidth}{@{\extracolsep{\fill}}llcccl@{}}
\toprule
Group & Name & Image $I$ & Question $q$ & State $z^\star$ & Requested / supervised output\\
\midrule
\multirow{4}{*}{Diagnostic} & Extract & Yes & Yes & No & $\hat z$ (\texttt{TASK\_STATE})\\
 & Solve & No & Yes & Yes & $\hat y^{\mathrm S}$ (\texttt{FINAL\_ANSWER})\\
 & Direct & Yes & Yes & No & $\hat y^{\mathrm D}$ (\texttt{FINAL\_ANSWER})\\
 & Zero-shot State-first & Yes & Yes & No & $\hat y$ (\texttt{FINAL\_ANSWER} only)\\
\midrule
\multirow{6}{*}{Training} & AO & Yes & Yes & No & $y^\star$ (\texttt{FINAL\_ANSWER})\\
 & FF & Yes & Yes & No & $\concat(e^\star,y^\star)$ (prose facts, \texttt{FINAL\_ANSWER})\\
 & SRT & Yes & Yes & No & $\concat(z^\star,y^\star)$ (\texttt{TASK\_STATE}, \texttt{FINAL\_ANSWER})\\
 & CoT & Yes & Yes & No & $\concat(e^\star,d^\star,y^\star)$ (facts, derivation, \texttt{FINAL\_ANSWER})\\
 & Placeholder-State & Yes & Yes & No & $\concat(z_\varnothing,y^\star)$ (\texttt{CHECK\_STATE: X}, \texttt{FINAL\_ANSWER})\\
 & State-after-Answer & Yes & Yes & No & $\concat(y^\star,z^\star)$ (\texttt{FINAL\_ANSWER}, \texttt{TASK\_STATE})\\
\bottomrule
\end{tabular*}
\end{table}

\subsection{Prompt Templates and Target Formats}
On pair-state ChartQA and CLEVR tasks, user messages begin with a domain prefix:
\begin{quote}\small\ttfamily\raggedright
ChartQA: Answer the chart question concisely. Do not explain your reasoning. Follow the requested output format exactly.\\[2pt]
CLEVR: Answer the CLEVR task concisely. Do not explain your reasoning. Follow the requested output format exactly.
\end{quote}
Here \texttt{\{answer-domain\}} is \texttt{FIRST|SECOND|EQUAL} for comparison, \texttt{<number>} for ChartQA arithmetic, and \texttt{<integer>} for CLEVR arithmetic. The three diagnostic interfaces use the following templates (shown for ChartQA, with CLEVR replacing \texttt{numeric value referenced first/second} by \texttt{answer to the first/second count query}):
\begin{quote}\small\ttfamily\raggedright
\textbf{Extract:}\\
\{prefix\} Full question: \{question\}\\
Extraction task: Do not answer the question. Read only the two numeric operands it asks about. Report them in question order as exactly one TASK\_STATE field and nothing else.\\
Required response format: TASK\_STATE: <value1>, <value2>\\[4pt]
\textbf{Solve:}\\
\{prefix\} Full question: \{question\}\\
The numeric value referenced first in the question is \{value1\}.\\
The numeric value referenced second in the question is \{value2\}.\\
Use these provided values to answer the question without an image.\\
Required response format: FINAL\_ANSWER: \{answer-domain\}\\[4pt]
\textbf{Direct (and AO):}\\
\{prefix\} \{question\}\\
Answer the question directly without reporting intermediate values.\\
Required response format: FINAL\_ANSWER: \{answer-domain\}
\end{quote}
In the two-call E$\to$S pipeline, the first call runs Extract and parses the predicted pair $(\hat a,\hat b)$, and the second call substitutes $(\hat a,\hat b)$ into the image-free Solve prompt without any fallback to $z^\star$. In image-conditioned Solve, the image is retained and the penultimate line of Solve reads \texttt{Use these provided values to answer the question.} Extract and Direct each receive the same complete question in a single call; the task-state components and object inventories illustrated in Figure~1 of the main paper are not additional prompts or text inputs.

\noindent\textbf{Training instructions.}
AO uses the Direct instruction. PS requests \texttt{CHECK\_STATE: X} before the answer, whereas SA requests the answer before the task state. Pair-state SRT asks the model to output both named fields in the requested order. On pair-state ChartQA, FF asks the model to write the two referenced values in question order as one sentence without a calculation. On Mixture, FF asks for the original visual facts in ordinary language, without presenting a computed result as directly observed. CoT asks the model to think through the question step by step before giving the final answer. FF and CoT share the same supervised fact span $e^\star$; only CoT adds the verified derivation $d^\star$. Table~\ref{tab:s-targets} illustrates the six response formats.

\noindent\textbf{Mixture state formats.}
Extract and SRT include the corresponding state schema in the prompt:
\begin{itemize}[leftmargin=*,topsep=2pt,itemsep=2pt]
\item \textbf{Multi-step arithmetic:} {\small\ttfamily \{"values":\allowbreak["number A",\allowbreak"number B",\allowbreak"number C"]\}} (three chart values in question order).
\item \textbf{Conditional lookup:} {\small\ttfamily \{"A":\allowbreak["value at P",\allowbreak"value at Q"],\allowbreak"B":[...],\allowbreak"C":[...]\}} (all six values without winner selection).
\item \textbf{Relational count ranking:} {\small\ttfamily \{"A":\allowbreak count\_A,\allowbreak"B":\allowbreak count\_B,\allowbreak"C":\allowbreak count\_C\}} (three integer counts in $A,B,C$ order).
\item \textbf{Color-set operations:} {\small\ttfamily \{"group1":\allowbreak["color",...],\allowbreak"group2":\allowbreak["color",...]\}} (distinct sorted colors per group).
\end{itemize}

\begin{table}[!htbp]
\centering
\small
\setlength{\tabcolsep}{6pt}
\caption{\textbf{Assistant target formats on a ChartQA sum question ($a=37$, $b=52$, $y^\star=89$).} Field names are literal and each displayed line corresponds to one output line.}
\label{tab:s-targets}
\begin{tabular*}{\textwidth}{@{\extracolsep{\fill}}ll@{}}
\toprule
Target & Supervised assistant response $R_c$\\
\midrule
AO & \texttt{FINAL\_ANSWER: 89}\\[2pt]
FF & \shortstack[l]{\texttt{The first value referenced in the question is 37, and the second is 52.}\\\texttt{FINAL\_ANSWER: 89}}\\[2pt]
SRT & \shortstack[l]{\texttt{TASK\_STATE: 37, 52}\\\texttt{FINAL\_ANSWER: 89}}\\[2pt]
CoT & \shortstack[l]{\texttt{The first value referenced in the question is 37, and the second is 52.}\\\texttt{Their sum is 37 + 52 = 89.}\\\texttt{FINAL\_ANSWER: 89}}\\[2pt]
Placeholder-State & \shortstack[l]{\texttt{CHECK\_STATE: X}\\\texttt{FINAL\_ANSWER: 89}}\\[2pt]
State-after-Answer & \shortstack[l]{\texttt{FINAL\_ANSWER: 89}\\\texttt{TASK\_STATE: 37, 52}}\\
\bottomrule
\end{tabular*}
\end{table}

\subsection{Canonicalization and Scoring Rules}
\label{sec:s-scoring}
StateLens results use strict canonicalization $\kappa(\cdot)$. Results on original ChartQA instead use the benchmark's relaxed accuracy, with a $5\%$ relative tolerance for numerical answers. Numeric strings undergo Unicode NFKC normalization, whitespace trimming, comma removal, optional trailing percent-sign removal on ChartQA without rescaling, and exact finite-decimal canonicalization (so \texttt{12.50} and \texttt{12.5} map to the same canonical decimal, while numerical approximations do not match). Comparison answers map to \texttt{FIRST}, \texttt{SECOND}, or \texttt{EQUAL}, ranking answers to \texttt{A}, \texttt{B}, or \texttt{C}, and color-set answers to sorted duplicate-free lowercase color sets (with \texttt{none} for the empty set). Under strict scoring, single-field interfaces require a well-formed \texttt{FINAL\_ANSWER} (or \texttt{TASK\_STATE}) field or a bare single-line value in the target domain, and two-field structured targets require both fields in the requested order. A malformed required field maps to an invalid symbol that matches no reference, whereas a well-formed but factually wrong intermediate field does not invalidate a correct \texttt{FINAL\_ANSWER}. Written facts in FF and CoT are parsed from their structured sentence templates, and JSON states in SRT are parsed and compared field by field against $z^\star$.

We also evaluate three scoring variants on fixed outputs: \emph{content-normalized Direct scoring} accepts an explicit \texttt{FINAL\_ANSWER} field or a valid final-line answer with surrounding text; \emph{order-relaxed Extract scoring} accepts swapped pairs on sum and absolute difference but not on ordered comparison; and \emph{content-only answer scoring} for trained models checks the final answer without requiring a valid auxiliary field.

\subsection{Prompt Templates for Parallax on Original ChartQA}
\label{sec:s-parallax-prompts}

Each route receives the original chart image and the unmodified question. The templates omit the common image input and denote the question text by \texttt{\{question\}}.

\newenvironment{parallaxprompt}{%
  \begin{list}{}{\setlength{\leftmargin}{1em}\setlength{\rightmargin}{0pt}%
    \setlength{\topsep}{3pt}\setlength{\partopsep}{0pt}\setlength{\parsep}{0pt}}%
  \item\small\raggedright
}{\end{list}}

\noindent\textbf{Direct (thinking disabled).}
The short-answer instruction follows the VLMEvalKit image VQA template~\cite{vlmevalkit2024}.
\begin{parallaxprompt}
\{question\}\\
Answer the question using a single word or phrase.
\end{parallaxprompt}

\noindent\textbf{Challenger.}
\begin{parallaxprompt}
Question: \{question\}\par\smallskip
Read the chart values or labels needed for this question, keeping each one associated with the correct category, series, and time. Briefly state those relevant observations, then perform only the required comparison or calculation. For a direct lookup, return the observed value. Do not describe the entire chart. Keep visually observed information distinct from values calculated from it.\par\smallskip
Give the requested answer as a single word, short phrase, or number. For a numerical answer, report the requested numerical magnitude without explanatory unit words or a percent sign. Keep the numerical scale requested by the question or used for the plotted quantity; do not silently convert a plotted percentage to a fraction or expand a chart-scaled value. If the question asks for a unit, label, or name itself, return that text rather than deleting it. State only what the chart and question support. End with exactly one line: The answer is X.
\end{parallaxprompt}

\noindent\textbf{Step-by-step CoT.}
We adapt the Factoid CoT prompt from ChartQAPro~\cite{masry2025chartqapro}.
\begin{parallaxprompt}
You are given a factoid question that you need to answer based on the provided image. You need to think step-by-step, but your final answer should be a single word, number, or phrase. If the question is unanswerable based on the information in the provided image, your answer should be unanswerable. Do not generate units. But if numerical units such as million, m, billion, B, or K are required, use the exact notation shown in the chart. If there are multiple final answers, put them in brackets using this format ['Answer1', 'Answer2']. Remember to think step-by-step and format the final answer in a separate sentence like "The answer is X".\\
Question: \{question\}
\end{parallaxprompt}

\noindent\textbf{Zero-shot CoT.}
We use the public Llama 3.2 Vision ChartQA prompt~\cite{llama2024vision}, with the image passed through the Qwen input interface.
\begin{parallaxprompt}
You are provided a chart image and will be asked a question. You have to think through your answer and provide a step-by-step solution. Once you have the solution, write the final answer in at most a few words at the end with the phrase "FINAL ANSWER:". The question is: \{question\}\\
Let's think step by step.
\end{parallaxprompt}

\noindent\textbf{Selector and Zero-shot judge.}
For the order with Challenger as Option B, the selector receives the following input:
\begin{parallaxprompt}
Question: \{question\}\par\smallskip
Select the answer better supported by the image and the actual question.\\
Option A: \{answer A\}\\
Option B: \{answer B\}\\
The options are unverified model predictions; neither is a reference answer. Read the relevant chart values and interpret the requested operation. Output only A or B.\par\smallskip
Model-generated draft for Option B (unverified):\\
{[Draft content withheld.]}\\
The draft may misread the chart or perform the wrong operation. Check its relevant values and calculation against the image and question rather than accepting its conclusion because it is detailed. Select between the unchanged options above. Output only A or B.
\end{parallaxprompt}
The header moves with the Challenger answer when the options are swapped. Parallax and Untrained scoring use the withheld-draft marker. Zero-shot judge and the trained full-draft variant replace the marker with the original draft. For the trained variants, keeping the rest of the prompt fixed isolates the effect of exposing the draft. The selector uses the A/B next-token scores; final-answer accuracy uses the relaxed ChartQA rule in Section~\ref{sec:s-scoring}.

\FloatBarrier
\section{Extended Diagnostic and Robustness Analyses}
\label{sec:s-diagnosis}

\subsection{Error Partition, Operations, and Mixture Families}
As shown in Table~\ref{tab:s-exs}, across all seven model--domain settings of the main-paper diagnosis, when Extract fails ($E=0$) the base model still succeeds on Solve ($S=1$) on the vast majority of questions, and simultaneous failures of both Extract and Solve ($E=0,S=0$) represent at most $9.7\%$ of Direct errors ($0.0\%$ on Qwen3.5). Because the two-call E$\to$S pipeline passes the canonicalized Extract prediction $\kappa(\hat z_i)$ into the deterministic Solve prompt, every question in $\Fcmp=\{i:E_i=1,S_i=1,D_i=0\}$ satisfies $\kappa(\hat z_i)=\kappa(z_i^\star)$ and therefore receives the same canonical state input as ground-truth Solve, so $\{i:E_i=1,S_i(\hat z_i)=1,D_i=0\}=\Fcmp$. Item-level E$\to$S evaluation logs on ChartQA directly verify $158/158$ (Qwen2.5), $268/268$ (Qwen3.5), $296/296$ (Molmo), $272/272$ (InternVL), and $234/234$ (MiniCPM) exact recoveries on $\Fcmp$. On the $33{,}723$-item CLEVR split, where visual count extraction is the primary bottleneck, $|\Jset|/N$ under the same deterministic handoff gives a lower bound of $17{,}916/33{,}723=53.1\%$ for Qwen2.5, exceeding Direct ($37.5\%$) by $15.6$ points. As shown in Table~\ref{tab:s-ops}, composition failures appear across all three ChartQA operations (comparison, absolute difference, and sum). Table~\ref{tab:s-qwen35-ops} further breaks down the seed-42 trained accuracies on Qwen3.5 across the three operations ($359$ test questions each), showing that SRT improves over AO, Placeholder-State, and State-after-Answer on comparison, difference, and sum alike.

\begin{table}[!htbp]
\centering
\small
\setlength{\tabcolsep}{3.5pt}
\caption{\textbf{Qwen3.5 accuracy (\%) by operation on ChartQA} ($359$ questions per operation, seed $42$).}
\label{tab:s-qwen35-ops}
\begin{tabular*}{\columnwidth}{@{\extracolsep{\fill}}lrrr@{}}
\toprule
Target & Comparison & Difference & Sum\\
\midrule
AO & 93.59 & 82.45 & 88.02\\
Placeholder-State & 94.99 & 83.57 & 85.52\\
State-after-Answer & 96.10 & 83.01 & 86.35\\
SRT & \textbf{99.44} & \textbf{93.04} & \textbf{92.48}\\
\bottomrule
\end{tabular*}
\end{table}

\begin{table}[!htbp]
\centering
\small
\setlength{\tabcolsep}{4.5pt}
\caption{\textbf{Joint Extract, Solve, and Direct outcomes across the seven settings in the main diagnostic table.} Entries are exact question counts. Within each Direct outcome, columns are indexed by $(E,S)$. Under $D=0$, the first two cells form $\Fext$, the third is $\Fsol$, and the fourth is $\Fcmp$.}
\label{tab:s-exs}
\begin{tabular*}{\textwidth}{@{\extracolsep{\fill}}llr*{8}{r}@{}}
\toprule
 & & & \multicolumn{4}{c}{$D=0$ (Direct incorrect)} & \multicolumn{4}{c}{$D=1$ (Direct correct)}\\
\cmidrule(lr){4-7}\cmidrule(l){8-11}
Data & Model & $N$ & $(0,0)$ & $(0,1)$ & $(1,0)$ & $(1,1)$ & $(0,0)$ & $(0,1)$ & $(1,0)$ & $(1,1)$\\
\midrule
ChartQA & Qwen2.5 & 1{,}077 & 38 & 106 & 91 & 158 & 1 & 59 & 12 & 612\\
ChartQA & Qwen3.5 & 1{,}077 & 0 & 97 & 5 & 268 & 0 & 31 & 0 & 676\\
ChartQA & Molmo & 1{,}077 & 13 & 133 & 10 & 296 & 9 & 109 & 29 & 478\\
ChartQA & InternVL & 1{,}077 & 2 & 74 & 12 & 272 & 0 & 29 & 12 & 676\\
ChartQA & MiniCPM & 1{,}077 & 3 & 165 & 12 & 234 & 0 & 78 & 6 & 579\\
\midrule
CLEVR & Qwen2.5 & 33{,}723 & 783 & 9{,}550 & 1{,}035 & 9{,}708 & 9 & 4{,}390 & 40 & 8{,}208\\
CLEVR & Molmo & 33{,}723 & 151 & 18{,}594 & 13 & 4{,}044 & 146 & 7{,}771 & 34 & 2{,}970\\
\bottomrule
\end{tabular*}
\end{table}

\begin{table}[!htbp]
\centering
\small
\setlength{\tabcolsep}{5pt}
\caption{\textbf{Composition failures by operation on ChartQA.} CFS and CFR are percentages.}
\label{tab:s-ops}
\begin{tabular*}{\columnwidth}{@{\extracolsep{\fill}}llrrr@{}}
\toprule
Model & Operation & CF & CFS & CFR\\
\midrule
\multirow{3}{*}{Qwen2.5} & Comparison & 53 & 29.0 & 26.4\\
 & Difference & 55 & 52.4 & 19.9\\
 & Sum & 50 & 47.6 & 17.1\\
\midrule
\multirow{3}{*}{Molmo} & Comparison & 60 & 75.0 & 24.8\\
 & Difference & 107 & 59.8 & 43.1\\
 & Sum & 129 & 66.8 & 45.4\\
\bottomrule
\end{tabular*}
\end{table}

Table~\ref{tab:s-mix-diag} reports the base-model diagnosis on each family of the four-family mixture as well as the pooled mixture totals referenced in the main paper. Over the $1{,}306$ mixture questions, Qwen2.5 achieves $|\Jset|=277$ joint successes and $154$ composition failures ($\cfr=55.6\%$), Qwen3.5 achieves $|\Jset|=501$ joint successes and $72$ composition failures ($\cfr=14.4\%$), and Molmo achieves $|\Jset|=105$ joint successes and $76$ composition failures ($\cfr=72.4\%$). Qwen2.5 exhibits composition failures on every family, Qwen3.5 concentrates its composition failures on relational count ranking ($47/62$, $\cfr=75.8\%$) and color-set operations ($25/221$, $\cfr=11.3\%$), and Molmo has no joint Extract--Solve successes on relational count ranking ($|\Jset|=0$, where CFR is undefined) and substantial CFR on the other three families.

\begin{table}[!htbp]
\centering
\small
\setlength{\tabcolsep}{4.5pt}
\caption{\textbf{Base-model diagnosis on the four-family mixture.} $|\Jset|$ counts questions with $E_i=S_i=1$, CF counts composition failures ($|\Fcmp|$), and 95\% CI gives exact two-sided Clopper--Pearson intervals for CFR (\%).}
\label{tab:s-mix-diag}
\begin{tabular*}{\textwidth}{@{\extracolsep{\fill}}llrrrrrc@{}}
\toprule
Model & Family & $N$ & $|\Jset|$ & CF & CFS (\%) & CFR (\%) & CFR 95\% CI\\
\midrule
\multirow{5}{*}{Qwen2.5} & Multi-step arithmetic & 291 & 26 & 25 & 9.2 & 96.2 & [80.4, 99.9]\\
 & Conditional lookup & 215 & 65 & 33 & 26.0 & 50.8 & [38.1, 63.4]\\
 & Relational count ranking & 400 & 35 & 28 & 10.9 & 80.0 & [63.1, 91.6]\\
 & Color-set operations & 400 & 151 & 68 & 31.5 & 45.0 & [36.9, 53.3]\\
\cmidrule(l){2-8}
 & All mixture families & 1{,}306 & 277 & 154 & 17.7 & 55.6 & [49.5, 61.5]\\
\midrule
\multirow{5}{*}{Qwen3.5} & Multi-step arithmetic & 291 & 100 & 0 & 0.0 & 0.0 & [0.0, 3.6]\\
 & Conditional lookup & 215 & 118 & 0 & 0.0 & 0.0 & [0.0, 3.1]\\
 & Relational count ranking & 400 & 62 & 47 & 14.4 & 75.8 & [63.3, 85.8]\\
 & Color-set operations & 400 & 221 & 25 & 37.9 & 11.3 & [7.5, 16.2]\\
\cmidrule(l){2-8}
 & All mixture families & 1{,}306 & 501 & 72 & 15.9 & 14.4 & [11.4, 17.8]\\
\midrule
\multirow{5}{*}{Molmo} & Multi-step arithmetic & 291 & 57 & 49 & 17.4 & 86.0 & [74.2, 93.7]\\
 & Conditional lookup & 215 & 4 & 2 & 1.1 & 50.0 & [6.8, 93.2]\\
 & Relational count ranking & 400 & 0 & 0 & 0.0 & -- & --\\
 & Color-set operations & 400 & 44 & 25 & 8.1 & 56.8 & [41.0, 71.7]\\
\cmidrule(l){2-8}
 & All mixture families & 1{,}306 & 105 & 76 & 6.7 & 72.4 & [62.8, 80.7]\\
\bottomrule
\end{tabular*}
\end{table}

\FloatBarrier
\subsection{Task-State Annotation on Human-Written ChartQA}
\label{sec:s-natural-esd}

\noindent\textbf{Question and state selection.}
We retain the original wording and reference answers of 443 human-written questions from the official ChartQA test split, one question per chart. Selection requires the visual facts needed by the question to be identifiable and expressible as named entries. Each reference state specifies the relevant labels, series, units, and values. If a question asks for an extreme or a comparison, the state contains the candidate entries needed to make that decision, rather than just the winning entry or the final answer. The reference states are checked against the original chart images before the Extract outputs are reviewed.

\noindent\textbf{Interfaces and evaluation.}
Extract receives the image and unchanged question and returns a list of named facts. Solve receives the question and reference facts without the image, whereas Direct receives the image and unchanged question without reference facts. Both answer interfaces permit intermediate calculations and reasoning and request a final \texttt{FINAL\_ANSWER} field. Extract is correct only when the response supplies all required facts, including the units and correct label--value bindings for each relevant candidate, regardless of JSON formatting; uncertain matches are counted as incorrect. Solve and Direct score only the final answer, with a 5\% relative tolerance for nonzero numerical answers and exact matching for zero, years, and categorical answers. This Direct interface differs from the short-answer protocol used for Parallax on all 2,500 original ChartQA questions.

\begin{table}[!htbp]
\centering
\small
\setlength{\tabcolsep}{5pt}
\caption{\textbf{Composition failures on human-written ChartQA questions.} E, S, and D are counts of correct outcomes. CFS uses Direct errors as its denominator; CFR uses joint Extract--Solve successes ($|\Jset|$).}
\label{tab:s-natural-esd}
\begin{tabular*}{\textwidth}{@{\extracolsep{\fill}}llrrrrrrcc@{}}
\toprule
Sample & Model & $N$ & E & S & D & $|\Jset|$ & CF & CFS (\%) & CFR (\%)\\
\midrule
Selected questions & Qwen2.5 & 443 & 245 & 364 & 356 & 201 & 20 & 23.0 & 10.0\\
 & Molmo & 443 & 120 & 330 & 360 & 91 & 11 & 13.3 & 12.1\\
\midrule
Source-checked subset & Qwen2.5 & 142 & 75 & 123 & 121 & 65 & 7 & 33.3 & 10.8\\
 & Molmo & 142 & 42 & 109 & 125 & 34 & 2 & 11.8 & 5.9\\
\bottomrule
\end{tabular*}
\end{table}

As shown in Table~\ref{tab:s-natural-esd}, Extract accuracy ($245/443$ for Qwen2.5 and $120/443$ for Molmo) is lower than Direct accuracy on this subset. For questions involving multiple series or candidate entries, Extract must recover every relevant binding, whereas Direct is judged on the answer alone. The stricter state requirement narrows the jointly solvable pool $\Jset$, as in CLEVR. Even so, composition failures account for $23.0\%$ and $13.3\%$ of Direct errors across the 443 questions for Qwen2.5 and Molmo, respectively. On a source-table-grounded subset of 142 questions where every reference entry is additionally tied to an explicit source-table cell and role, composition failures continue to account for $33.3\%$ ($7/21$) and $11.8\%$ ($2/17$) of Direct errors for Qwen2.5 and Molmo, respectively.

\FloatBarrier
\subsection{Scoring, Wording, and Interface Robustness}
\noindent\textbf{Content-normalized Direct scoring and prompt paraphrases.}
As shown in Table~\ref{tab:s-content}, relaxing Direct scoring to accept answers embedded in extra text raises Molmo's ChartQA Direct accuracy from $58.0\%$ to $61.6\%$ and Qwen2.5's CLEVR Direct accuracy from $37.5\%$ to $38.6\%$, while leaving CFS above $45\%$ and CFR above $28\%$ across all tested settings. Repeating all three interfaces with two independent prompt paraphrases (Paraphrase A and Paraphrase B) similarly preserves CFR between $30.2\%$ and $34.5\%$ for Molmo and between $27.6\%$ and $28.8\%$ for MiniCPM, as shown in Table~\ref{tab:s-prompt}.

\begin{table}[!htbp]
\centering
\small
\setlength{\tabcolsep}{3pt}
\caption{\textbf{Sensitivity to content-normalized Direct scoring.} Extract and Solve outcomes are unchanged, and values are percentages.}
\label{tab:s-content}
\begin{tabular*}{\columnwidth}{@{\extracolsep{\fill}}llrrrr@{}}
\toprule
Data & Model & Direct & Direct (norm.) & CFS & CFR\\
\midrule
ChartQA & Molmo & 58.0 & 61.6 & 63.5 & 34.0\\
ChartQA & MiniCPM & 61.6 & 61.6 & 56.5 & 28.8\\
CLEVR & Qwen2.5 & 37.5 & 38.6 & 45.5 & 52.6\\
\bottomrule
\end{tabular*}
\end{table}

\begin{table}[!htbp]
\centering
\small
\setlength{\tabcolsep}{4pt}
\caption{\textbf{CFR (\%) across three prompt wordings on ChartQA} under content-normalized scoring.}
\label{tab:s-prompt}
\begin{tabular*}{\columnwidth}{@{\extracolsep{\fill}}lrrr@{}}
\toprule
Model & Default wording & Paraphrase A & Paraphrase B\\
\midrule
Molmo & 34.0 & 34.5 & 30.2\\
MiniCPM & 28.8 & 28.8 & 27.6\\
\bottomrule
\end{tabular*}
\end{table}

\noindent\textbf{Order-relaxed Extract scoring and image-conditioned Solve.}
As shown in Table~\ref{tab:s-order}, accepting reversed operand pairs $(b,a)$ on sum and absolute-difference questions changes CFR by at most $0.6$ points across all settings. As shown in Table~\ref{tab:s-image-solve}, retaining the chart image during Solve keeps accuracy close to image-free Solve ($85.4\%$ vs.\ $86.8\%$ for Qwen2.5 and $92.1\%$ vs.\ $94.3\%$ for Molmo). When $z^\star$ is replaced by an answer-changing counterfactual state $z_{\mathrm{cf}}$ while the original image remains visible, Qwen2.5 and Molmo follow the counterfactual answer on $45.7\%$ and $38.4\%$ of questions, respectively. Thus, on a substantial fraction of questions Solve follows the supplied task state even when it contradicts the image.

\begin{table}[!htbp]
\centering
\small
\setlength{\tabcolsep}{5pt}
\caption{\textbf{Order-relaxed Extract scoring on sum and absolute-difference questions.} Reversed counts Extract outputs whose two values match $z^\star$ in swapped order. Arrows show strict$\to$relaxed values.}
\label{tab:s-order}
\begin{tabular*}{\textwidth}{@{\extracolsep{\fill}}llrrrr@{}}
\toprule
Data & Model & Reversed & $|\Jset|$ & CF & CFR (\%)\\
\midrule
ChartQA & Qwen2.5 & 10 & $770\to780$ & $158\to158$ & $20.5\to20.3$\\
ChartQA & Molmo & 31 & $774\to805$ & $296\to303$ & $38.2\to37.6$\\
ChartQA & InternVL & 3 & $948\to951$ & $272\to272$ & $28.7\to28.6$\\
ChartQA & MiniCPM & 10 & $813\to823$ & $234\to236$ & $28.8\to28.7$\\
CLEVR & Qwen2.5 & 212 & $17{,}916\to18{,}128$ & $9{,}708\to9{,}793$ & $54.2\to54.0$\\
\bottomrule
\end{tabular*}
\end{table}

\begin{table}[!htbp]
\centering
\small
\setlength{\tabcolsep}{3.5pt}
\caption{\textbf{Solve accuracy (\%) on ChartQA with and without the image.} The counterfactual column reports agreement (\%) with the counterfactual target answer $y^\star_{\mathrm{cf}}$ after replacing $z^\star$ with an answer-changing state $z_{\mathrm{cf}}$ while keeping the image visible.}
\label{tab:s-image-solve}
\begin{tabular*}{\columnwidth}{@{\extracolsep{\fill}}lrrr@{}}
\toprule
Model & Solve (no $I$) & Solve ($I + z^\star$) & CF Match ($I + z_{\mathrm{cf}}$)\\
\midrule
Qwen2.5 & 86.8 & 85.4 & 45.7\\
Molmo & 94.3 & 92.1 & 38.4\\
\bottomrule
\end{tabular*}
\end{table}

\FloatBarrier
\section{Extended Supervision and Fact-Answer Analyses}
\label{sec:s-training}

\subsection{Exact Seed-42 Counts, Token Lengths, and Paired Comparisons}

\begin{table}[!htbp]
\centering
\small
\setlength{\tabcolsep}{4pt}
\renewcommand{\arraystretch}{0.87}
\caption{\textbf{Exact correct-answer counts and mean output tokens (in parentheses) in the seed-42 four-target comparison.} Every target in a row is evaluated on the same $N$ test questions. Mean output tokens are reported in parentheses where natively logged.}
\label{tab:s-exact}
\begin{tabular*}{\textwidth}{@{\extracolsep{\fill}}lllrrrrr@{}}
\toprule
Data & Family & Model & $N$ & AO & SRT & FF & CoT\\
\midrule
\multirow{5}{*}{ChartQA} & Pair-state & Qwen2.5 & 1{,}077 & 927 (7.97) & 991 (20.85) & \textbf{998} (29.87) & 996 (48.67)\\
 & Pair-state & Qwen3.5 & 1{,}077 & 948 (7.99) & \textbf{1{,}023} (20.85) & 1{,}017 (30.85) & 1{,}015 (50.68)\\
 & Pair-state & Molmo & 1{,}077 & 839 (7.30) & \textbf{989} (17.83) & 982 (26.83) & 978 (42.62)\\
 & Pair-state & InternVL & 1{,}077 & 934 (7.97) & 1{,}009 (20.88) & 1{,}010 (29.87) & \textbf{1{,}011} (48.65)\\
 & Pair-state & MiniCPM & 1{,}077 & 787 & 923 & \textbf{924} (29.85) & 920 (48.63)\\
\midrule
\multirow{5}{*}{Mixture} & Multi-step arithmetic & Qwen2.5 & 291 & 94 & 197 & 199 & \textbf{243}\\
 & Conditional lookup & Qwen2.5 & 215 & 154 & 183 & 188 & \textbf{191}\\
 & Relational count ranking & Qwen2.5 & 400 & 351 & 386 & \textbf{391} & 389\\
 & Color-set operations & Qwen2.5 & 400 & 318 & 394 & \textbf{398} & 395\\
 & All mixture families & Qwen2.5 & 1{,}306 & 917 (7.40) & 1{,}160 (28.07) & 1{,}176 (31.82) & \textbf{1{,}218} (56.19)\\
\midrule
\multirow{5}{*}{Mixture} & Multi-step arithmetic & Qwen3.5 & 291 & 148 & 246 & 254 & \textbf{259}\\
 & Conditional lookup & Qwen3.5 & 215 & 172 & 197 & \textbf{199} & \textbf{199}\\
 & Relational count ranking & Qwen3.5 & 400 & 364 & \textbf{390} & 382 & 388\\
 & Color-set operations & Qwen3.5 & 400 & 348 & 394 & \textbf{399} & 388\\
 & All mixture families & Qwen3.5 & 1{,}306 & 1{,}032 (7.39) & 1{,}227 (28.98) & \textbf{1{,}234} (32.78) & \textbf{1{,}234} (58.52)\\
\midrule
\multirow{5}{*}{Mixture} & Multi-step arithmetic & Molmo & 291 & 66 & 200 & 200 & \textbf{241}\\
 & Conditional lookup & Molmo & 215 & 145 & 181 & 182 & \textbf{186}\\
 & Relational count ranking & Molmo & 400 & 154 & 320 & \textbf{326} & 320\\
 & Color-set operations & Molmo & 400 & 294 & 364 & \textbf{384} & 380\\
 & All mixture families & Molmo & 1{,}306 & 659 (6.87) & 1{,}065 (25.64) & 1{,}092 (29.46) & \textbf{1{,}127} (51.95)\\
\bottomrule
\end{tabular*}
\end{table}

Table~\ref{tab:s-exact} reports the exact correct-answer numerators and mean generated token lengths underlying the seed-42 comparison of AO, SRT, FF, and CoT across ChartQA (for all five VLMs) and all four families of the mixture (for Qwen2.5, Qwen3.5, and Molmo). Table~\ref{tab:s-paired} gives all $43$ paired question-level win/loss contrasts and 95\% image-cluster bootstrap confidence intervals ($20{,}000$ resamples of the test charts or scenes) across all five models on ChartQA and all three models on the mixture. On ChartQA, the reported 95\% intervals for SRT versus FF and SRT versus CoT include zero across all five models, while all three targets exceed AO by $+5.94$ to $+13.93$ points. On the mixture, all three evidence-writing targets exceed AO by $+14.93$ to $+35.83$ points across the three models. CoT exceeds FF by $+3.22$ points ($[1.93,4.55]$) on Qwen2.5 and $+2.68$ points ($[1.08,4.25]$) on Molmo, and matches FF overall on Qwen3.5 ($1{,}234$ correct each, with $24$ unique wins per arm and 95\% CI $[-1.07,+1.07]$), while exceeding SRT on multi-step arithmetic across all three models ($+15.81$ points on Qwen2.5, $+4.47$ points on Qwen3.5, and $+14.09$ points on Molmo). Where token counts are available, CoT generates roughly twice as many tokens as SRT on average. Its accuracy gains over FF and SRT are concentrated in multi-step arithmetic rather than spread across task families, indicating that the gains primarily reflect explicit intermediate calculation steps rather than longer responses alone.

\FloatBarrier
\begingroup
\small
\setlength{\tabcolsep}{5pt}
\renewcommand{\arraystretch}{0.92}
\setlength{\LTpre}{6pt}
\setlength{\LTpost}{6pt}
\begin{longtable}{@{\extracolsep{\fill}}lllrrrrc@{}}
\caption{\textbf{All 43 paired question-level comparisons and 95\% image-cluster bootstrap intervals ($20{,}000$ resamples, seed 42).} For each contrast $B - A$, ``$A$ only'' and ``$B$ only'' count questions answered right by exclusively $A$ or exclusively $B$, and $\Delta = \mathrm{Acc}(B) - \mathrm{Acc}(A)$ in percentage points.}
\label{tab:s-paired}\\
\toprule
Data & Model & Contrast ($B - A$) & $N$ & $A$ only & $B$ only & $\Delta$ (pp) & 95\% cluster CI (pp)\\
\midrule
\endfirsthead
\multicolumn{8}{@{}l}{\textbf{Table \thetable.} Continued from the preceding page.}\\[3pt]
\toprule
Data & Model & Contrast ($B - A$) & $N$ & $A$ only & $B$ only & $\Delta$ (pp) & 95\% cluster CI (pp)\\
\midrule
\endhead
\midrule
\multicolumn{8}{r@{}}{\footnotesize Continued on the next page.}\\
\endfoot
\bottomrule
\endlastfoot
ChartQA & Qwen2.5 & $\mathrm{FF} - \mathrm{AO}$ & 1{,}077 & 8 & 79 & $+6.59$ & $[+4.55, +8.73]$\\*
 & & $\mathrm{SRT} - \mathrm{AO}$ & 1{,}077 & 12 & 76 & $+5.94$ & $[+3.81, +8.17]$\\*
 & & $\mathrm{CoT} - \mathrm{AO}$ & 1{,}077 & 11 & 80 & $+6.41$ & $[+4.27, +8.64]$\\*
 & & $\mathrm{SRT} - \mathrm{FF}$ & 1{,}077 & 11 & 4 & $-0.65$ & $[-1.58, +0.09]$\\*
 & & $\mathrm{SRT} - \mathrm{CoT}$ & 1{,}077 & 16 & 11 & $-0.46$ & $[-1.67, +0.74]$\\
\midrule
ChartQA  & Qwen3.5 & $\mathrm{FF} - \mathrm{AO}$ & 1{,}077 & 8 & 77 & $+6.41$ & $[+4.46, +8.45]$\\*
 & & $\mathrm{SRT} - \mathrm{AO}$ & 1{,}077 & 5 & 80 & $+6.96$ & $[+5.11, +9.01]$\\*
 & & $\mathrm{CoT} - \mathrm{AO}$ & 1{,}077 & 9 & 76 & $+6.22$ & $[+4.27, +8.26]$\\*
 & & $\mathrm{SRT} - \mathrm{FF}$ & 1{,}077 & 5 & 11 & $+0.56$ & $[-0.37, +1.58]$\\*
 & & $\mathrm{SRT} - \mathrm{CoT}$ & 1{,}077 & 5 & 13 & $+0.74$ & $[-0.19, +1.86]$\\
\midrule
ChartQA  & Molmo & $\mathrm{FF} - \mathrm{AO}$ & 1{,}077 & 17 & 160 & $+13.28$ & $[+10.40, +16.25]$\\*
 & & $\mathrm{SRT} - \mathrm{AO}$ & 1{,}077 & 8 & 158 & $+13.93$ & $[+11.14, +16.81]$\\*
 & & $\mathrm{CoT} - \mathrm{AO}$ & 1{,}077 & 17 & 156 & $+12.91$ & $[+10.03, +15.88]$\\*
 & & $\mathrm{SRT} - \mathrm{FF}$ & 1{,}077 & 15 & 22 & $+0.65$ & $[-0.84, +2.14]$\\*
 & & $\mathrm{SRT} - \mathrm{CoT}$ & 1{,}077 & 15 & 26 & $+1.02$ & $[-0.46, +2.51]$\\
\midrule
ChartQA  & InternVL & $\mathrm{FF} - \mathrm{AO}$ & 1{,}077 & 6 & 82 & $+7.06$ & $[+5.11, +9.10]$\\*
 & & $\mathrm{SRT} - \mathrm{AO}$ & 1{,}077 & 10 & 85 & $+6.96$ & $[+4.92, +9.10]$\\*
 & & $\mathrm{CoT} - \mathrm{AO}$ & 1{,}077 & 8 & 85 & $+7.15$ & $[+5.11, +9.19]$\\*
 & & $\mathrm{SRT} - \mathrm{FF}$ & 1{,}077 & 9 & 8 & $-0.09$ & $[-1.02, +0.74]$\\*
 & & $\mathrm{SRT} - \mathrm{CoT}$ & 1{,}077 & 7 & 5 & $-0.19$ & $[-1.02, +0.56]$\\
\midrule
ChartQA  & MiniCPM & $\mathrm{FF} - \mathrm{AO}$ & 1{,}077 & 24 & 161 & $+12.72$ & $[+9.66, +15.88]$\\*
 & & $\mathrm{SRT} - \mathrm{AO}$ & 1{,}077 & 24 & 160 & $+12.63$ & $[+9.66, +15.69]$\\*
 & & $\mathrm{CoT} - \mathrm{AO}$ & 1{,}077 & 25 & 158 & $+12.35$ & $[+9.29, +15.41]$\\*
 & & $\mathrm{SRT} - \mathrm{FF}$ & 1{,}077 & 33 & 32 & $-0.09$ & $[-2.14, +1.95]$\\*
 & & $\mathrm{SRT} - \mathrm{CoT}$ & 1{,}077 & 27 & 30 & $+0.28$ & $[-1.58, +2.04]$\\
\midrule
Mixture & Qwen2.5 & $\mathrm{FF} - \mathrm{AO}$ & 1{,}306 & 21 & 280 & $+19.83$ & $[+17.44, +22.21]$\\*
 & & $\mathrm{SRT} - \mathrm{AO}$ & 1{,}306 & 32 & 275 & $+18.61$ & $[+16.19, +21.04]$\\*
 & & $\mathrm{CoT} - \mathrm{AO}$ & 1{,}306 & 13 & 314 & $+23.05$ & $[+20.67, +25.45]$\\*
 & & $\mathrm{SRT} - \mathrm{FF}$ & 1{,}306 & 41 & 25 & $-1.23$ & $[-2.50, +0.00]$\\*
 & & $\mathrm{CoT} - \mathrm{FF}$ & 1{,}306 & 15 & 57 & $+3.22$ & $[+1.93, +4.55]$\\*
 & & $\mathrm{CoT} - \mathrm{SRT}$ & 1{,}306 & 18 & 76 & $+4.44$ & $[+2.98, +5.93]$\\
\midrule
Mixture  & Qwen3.5 & $\mathrm{FF} - \mathrm{AO}$ & 1{,}306 & 20 & 222 & $+15.47$ & $[+13.31, +17.66]$\\*
 & & $\mathrm{SRT} - \mathrm{AO}$ & 1{,}306 & 19 & 214 & $+14.93$ & $[+12.85, +17.06]$\\*
 & & $\mathrm{CoT} - \mathrm{AO}$ & 1{,}306 & 18 & 220 & $+15.47$ & $[+13.35, +17.62]$\\*
 & & $\mathrm{SRT} - \mathrm{FF}$ & 1{,}306 & 30 & 23 & $-0.54$ & $[-1.62, +0.53]$\\*
 & & $\mathrm{CoT} - \mathrm{FF}$ & 1{,}306 & 24 & 24 & $0.00$ & $[-1.07, +1.07]$\\*
 & & $\mathrm{CoT} - \mathrm{SRT}$ & 1{,}306 & 25 & 32 & $+0.54$ & $[-0.54, +1.62]$\\
\midrule
Mixture  & Molmo & $\mathrm{FF} - \mathrm{AO}$ & 1{,}306 & 42 & 475 & $+33.15$ & $[+30.24, +36.05]$\\*
 & & $\mathrm{SRT} - \mathrm{AO}$ & 1{,}306 & 53 & 459 & $+31.09$ & $[+28.11, +34.05]$\\*
 & & $\mathrm{CoT} - \mathrm{AO}$ & 1{,}306 & 44 & 512 & $+35.83$ & $[+32.85, +38.73]$\\*
 & & $\mathrm{SRT} - \mathrm{FF}$ & 1{,}306 & 72 & 45 & $-2.07$ & $[-3.70, -0.46]$\\*
 & & $\mathrm{CoT} - \mathrm{FF}$ & 1{,}306 & 39 & 74 & $+2.68$ & $[+1.08, +4.25]$\\*
 & & $\mathrm{CoT} - \mathrm{SRT}$ & 1{,}306 & 29 & 91 & $+4.75$ & $[+3.13, +6.37]$\\
\end{longtable}
\endgroup

\FloatBarrier
\subsection{Additional Seeds and Prompting Baselines}
To verify stability across random initializations and data orderings, Table~\ref{tab:s-multiseed} reports three-seed means and sample standard deviations (seeds $13$, $42$, and $87$) on ChartQA for AO, Placeholder-State, State-after-Answer, and SRT, alongside CLEVR results. Across all four multi-seed ChartQA models, standard deviations of SRT are at most $0.6$ points, and SRT exceeds AO, Placeholder-State, and State-after-Answer by $5.9$--$15.8$ points. As shown in Table~\ref{tab:s-content-train}, rescoring the same checkpoints under content-only answer scoring leaves every margin unchanged to within $0.3$ points, confirming that the advantage of writing the state before the answer does not stem from two-field format compliance.

\begin{table}[!htbp]
\centering
\small
\setlength{\tabcolsep}{4.5pt}
\caption{\textbf{Multi-seed comparison of state content and position.} ChartQA entries report $\text{mean}\pm\text{sd}$ (\%) across seeds $13$, $42$, and $87$, and CLEVR entries use seed $42$.}
\label{tab:s-multiseed}
\begin{tabular*}{\textwidth}{@{\extracolsep{\fill}}llrrrrr@{}}
\toprule
Data & Model & Direct (base) & AO & Placeholder-State & State-after-Answer & SRT\\
\midrule
\multirow{4}{*}{ChartQA} & Qwen2.5 & 63.5 & $\ms{85.1}{0.8}$ & $\ms{85.8}{0.7}$ & $\ms{85.1}{0.6}$ & $\mathbf{\ms{91.7}{0.4}}$\\
 & Molmo & 58.0 & $\ms{77.3}{0.6}$ & $\ms{75.7}{2.4}$ & $\ms{79.4}{1.0}$ & $\mathbf{\ms{91.5}{0.6}}$\\
 & InternVL & 66.6 & $\ms{87.7}{0.8}$ & $\ms{86.2}{0.8}$ & $\ms{87.4}{1.2}$ & $\mathbf{\ms{94.0}{0.4}}$\\
 & MiniCPM & 61.6 & $\ms{72.8}{1.3}$ & $\ms{72.3}{1.7}$ & $\ms{73.0}{1.0}$ & $\mathbf{\ms{86.0}{0.4}}$\\
\midrule
\multirow{2}{*}{CLEVR} & Qwen2.5 & 37.5 & 93.7 & 92.9 & 93.5 & \textbf{95.4}\\
 & Molmo & 32.4 & 87.7 & 86.8 & 90.6 & \textbf{92.0}\\
\bottomrule
\end{tabular*}
\end{table}

\begin{table}[!htbp]
\centering
\small
\setlength{\tabcolsep}{4.5pt}
\caption{\textbf{Final-answer accuracy (\%) under content-only scoring} (auxiliary field validity ignored, three-seed means on ChartQA).}
\label{tab:s-content-train}
\begin{tabular*}{\textwidth}{@{\extracolsep{\fill}}llrrrrrr@{}}
\toprule
Data & Model & AO & Placeholder-State & State-after-Answer & SRT & SRT$-$Placeholder & SRT$-$After\\
\midrule
ChartQA & Qwen2.5 & 85.1 & 85.8 & 85.1 & \textbf{91.7} & $+5.9$ & $+6.7$\\
ChartQA & Molmo & 77.3 & 75.7 & 79.7 & \textbf{91.6} & $+15.9$ & $+11.9$\\
ChartQA & InternVL & 87.7 & 86.2 & 87.4 & \textbf{94.0} & $+7.8$ & $+6.6$\\
ChartQA & MiniCPM & 72.8 & 72.3 & 73.0 & \textbf{86.0} & $+13.7$ & $+13.0$\\
CLEVR & Qwen2.5 & 93.7 & 92.9 & 93.5 & \textbf{95.4} & $+2.5$ & $+1.9$\\
\bottomrule
\end{tabular*}
\end{table}

Table~\ref{tab:s-zeroshot} tests whether directing the base model to attend to the two task-relevant values helps under the same answer-only output contract, without parameter updates. This control changes the instruction but does not expose any intermediate state:
\begin{quote}\small\ttfamily\raggedright
\{prefix\} \{question\}\\
Answer the question using the image. First identify the two task-relevant values in the order they are referenced in the question. Use those values to solve the question. Do not report the intermediate values or your reasoning. Return only: FINAL\_ANSWER: \{answer-domain\}
\end{quote}
This prompt is distinct from both the two-call E$\to$S pipeline and single-response generation with visible intermediate text. Under this answer-only instruction, accuracy stays below Direct on four models and increases by $1.0$ point on Qwen3.5. InternVL returns an invalid answer-only response on $78.8\%$ of the questions.

\begin{table}[!htbp]
\centering
\small
\setlength{\tabcolsep}{6pt}
\caption{\textbf{Zero-shot state-first prompting vs.\ E$\to$S and SRT on ChartQA} (accuracy and invalid rate in \%, with SRT at seed 42).}
\label{tab:s-zeroshot}
\begin{tabular*}{\textwidth}{@{\extracolsep{\fill}}lrrrrr@{}}
\toprule
Model & Direct & Zero-shot & Invalid & E$\to$S & SRT\\
\midrule
Qwen2.5 & 63.5 & 59.1 & 0.1 & 75.1 & \textbf{92.0}\\
Qwen3.5 & 65.6 & 66.7 & 0.1 & 90.6 & \textbf{95.0}\\
Molmo & 58.0 & 53.4 & 5.7 & 79.7 & \textbf{91.8}\\
InternVL & 66.6 & 11.6 & 78.8 & 90.7 & \textbf{93.7}\\
MiniCPM & 61.6 & 53.9 & 13.6 & 82.1 & \textbf{85.7}\\
\bottomrule
\end{tabular*}
\end{table}

\FloatBarrier
\subsection{CoT Gains Across Base Diagnostic Strata}
For the eight settings with base $(E,S,D)$ diagnoses, Table~\ref{tab:s-cot-srt-partition} decomposes the net change in correct answers from SRT to CoT across fixed, disjoint strata. On Mixture, other base Direct errors account for 45 of the 58 net additional correct answers on Qwen2.5 and 60 of the 62 on Molmo, where each entry reports the net difference in correct predictions (CoT-only correct minus SRT-only correct) within the corresponding base stratum.

\begin{table}[!htbp]
\centering
\small
\setlength{\tabcolsep}{3pt}
\caption{\textbf{CoT minus SRT by base diagnostic stratum.} Entries are signed differences in correct counts.}
\label{tab:s-cot-srt-partition}
\begin{tabular}{@{}llrrrr@{}}
\toprule
Data & Model & Total & $\Fcmp^{(0)}$ & $\mathcal G^{(0)}$ & Other errors \\
\midrule
ChartQA & Qwen2.5 & $+5$ & $+3$ & $0$ & $+2$ \\
 & Molmo & $-11$ & $0$ & $-2$ & $-9$ \\
 & Qwen3.5 & $-8$ & $-3$ & $-1$ & $-4$ \\
 & InternVL & $+2$ & $0$ & $+3$ & $-1$ \\
 & MiniCPM & $-3$ & $+6$ & $0$ & $-9$ \\
Mixture & Qwen2.5 & $+58$ & $+5$ & $+8$ & $+45$ \\
 & Qwen3.5 & $+7$ & $-2$ & $+13$ & $-4$ \\
 & Molmo & $+62$ & $0$ & $+2$ & $+60$ \\
\bottomrule
\end{tabular}
\end{table}

\subsection{State Realization and State--Answer Contingency}
\label{sec:s-state-contingency}
Table~\ref{tab:s-state-repair} reports exact state accuracy ($\kappa(\hat z)=\kappa(z^\star)$), state--answer consistency ($\kappa(\hat y)=\kappa(\phi_o(\hat z))$), composition-failure repair on $\Fcmp^{(0)}$, and preservation on $\mathcal G^{(0)}$ across the three-seed ChartQA and CLEVR runs. In every reported pair-state setting, the written answer agrees with the result of applying $\phi_o$ to the written state $\hat z$ in at least $99.2\%$ of cases. On the four mixture families (Table~\ref{tab:s-mix-state}), consistency is similarly at least $96.2\%$ on conditional lookup, relational count ranking, and color-set operations, whereas on multi-step arithmetic it drops to $81.8\%$ for Qwen2.5 and $83.7\%$ for Molmo because chaining two operations without a written derivation remains error-prone even after the three operands are written correctly.

\begin{table}[!htbp]
\centering
\small
\setlength{\tabcolsep}{4.5pt}
\caption{\textbf{State realization, composition-failure repair, and preservation} (percentages, $\text{mean}\pm\text{sd}$ over three seeds on ChartQA except single-seed Qwen3.5 and CLEVR).}
\label{tab:s-state-repair}
\begin{tabular*}{\textwidth}{@{\extracolsep{\fill}}llrrrrr@{}}
\toprule
 & & \multicolumn{2}{c}{SRT state metrics} & \multicolumn{2}{c}{Repaired in $\Fcmp^{(0)}$} & \multirow{2}{*}{SRT preserved in $\mathcal G^{(0)}$}\\
\cmidrule(lr){3-4}\cmidrule(lr){5-6}
Data & Model & State Acc. & State--Answer Cons. & AO & SRT & \\
\midrule
\multirow{5}{*}{ChartQA} & Qwen2.5 & $\ms{89.1}{0.2}$ & $\ms{99.2}{0.5}$ & $\ms{73.0}{4.5}$ & $\mathbf{\ms{94.3}{1.1}}$ & $\ms{98.8}{0.1}$\\
 & Qwen3.5 & 92.4 & 100.0 & 82.5 & \textbf{99.6} & 99.6\\
 & Molmo & $\ms{88.1}{0.8}$ & $\ms{99.3}{0.1}$ & $\ms{71.4}{0.2}$ & $\mathbf{\ms{98.1}{1.4}}$ & $\ms{98.5}{0.6}$\\
 & InternVL & $\ms{91.7}{0.6}$ & $\ms{99.8}{0.2}$ & $\ms{82.4}{1.7}$ & $\mathbf{\ms{97.5}{0.4}}$ & $\ms{99.2}{0.3}$\\
 & MiniCPM & $\ms{81.2}{1.0}$ & $\ms{99.8}{0.1}$ & $\ms{58.3}{3.0}$ & $\mathbf{\ms{92.5}{2.4}}$ & $\ms{97.1}{0.1}$\\
\midrule
\multirow{2}{*}{CLEVR} & Qwen2.5 & 93.9 & 100.0 & 95.3 & \textbf{96.8} & 97.0\\
 & Molmo & 87.9 & 100.0 & 93.9 & \textbf{95.9} & 95.3\\
\bottomrule
\end{tabular*}
\end{table}

\begin{table}[!htbp]
\centering
\small
\setlength{\tabcolsep}{6pt}
\caption{\textbf{SRT state accuracy and state--answer consistency (\%) on the four mixture families} (seed $42$).}
\label{tab:s-mix-state}
\begin{tabular*}{\textwidth}{@{\extracolsep{\fill}}llrr@{}}
\toprule
Model & Family & State Acc. & S--A Cons.\\
\midrule
\multirow{4}{*}{Qwen2.5} & Multi-step arithmetic & 84.2 & 81.8\\
 & Conditional lookup & 79.5 & 96.2\\
 & Relational count ranking & 78.5 & 100.0\\
 & Color-set operations & 97.5 & 99.3\\
\midrule
\multirow{4}{*}{Molmo} & Multi-step arithmetic & 80.8 & 83.7\\
 & Conditional lookup & 78.6 & 99.1\\
 & Relational count ranking & 40.0 & 100.0\\
 & Color-set operations & 87.5 & 98.8\\
\bottomrule
\end{tabular*}
\end{table}

Table~\ref{tab:s-fact-acc} separates exact fact accuracy over all questions from the fraction of correct answers accompanied by incorrect facts. The latter divides the last column by the total number of correct answers. For example, Qwen3.5 FF on Mixture produces 133 answers with wrong facts among its 1,234 correct answers, giving 10.8\%. On Mixture, CoT has lower fact accuracy than FF on Qwen2.5 (84.0\% vs.\ 86.0\%) and Molmo (71.2\% vs.\ 74.0\%), even though it answers more questions correctly. This is consistent with the measured gains in downstream arithmetic rather than improved extraction across the dataset.

\FloatBarrier
\begingroup
\small
\setlength{\tabcolsep}{4.5pt}
\renewcommand{\arraystretch}{0.91}
\setlength{\LTpre}{6pt}
\setlength{\LTpost}{6pt}
\begin{longtable}{@{\extracolsep{\fill}}llrrrrrr@{}}
\caption{\textbf{Written-fact accuracy and joint fact--answer outcomes across FF, SRT, and CoT (seed 42).} Fact correct counts exact agreement with the reference task state. For Molmo CoT on Mixture, 2 additional correct answers have unparseable fact prefixes (1,127 correct answers in total).}
\label{tab:s-fact-acc}\\
\toprule
Data & Model & Target & $N$ & Fact correct & Fact Acc. (\%) & Fact right \& Ans. right & Fact wrong \& Ans. right\\
\midrule
\endfirsthead
\multicolumn{8}{@{}l}{\textbf{Table \thetable.} Continued from the preceding page.}\\[3pt]
\toprule
Data & Model & Target & $N$ & Fact correct & Fact Acc. (\%) & Fact right \& Ans. right & Fact wrong \& Ans. right\\
\midrule
\endhead
\midrule
\multicolumn{8}{r@{}}{\footnotesize Continued on the next page.}\\
\endfoot
\bottomrule
\endlastfoot
ChartQA & Qwen2.5 & FF & 1{,}077 & 969 & 89.97 & 965 & 33\\*
 & & SRT & 1{,}077 & 962 & 89.32 & 955 & 36\\*
 & & CoT & 1{,}077 & 968 & 89.88 & 965 & 31\\
\midrule
ChartQA  & Qwen3.5 & FF & 1{,}077 & 990 & 91.92 & 990 & 27\\*
 & & SRT & 1{,}077 & 995 & 92.39 & 995 & 28\\*
 & & CoT & 1{,}077 & 989 & 91.83 & 988 & 27\\
\midrule
ChartQA  & Molmo & FF & 1{,}077 & 948 & 88.02 & 945 & 37\\*
 & & SRT & 1{,}077 & 955 & 88.67 & 952 & 37\\*
 & & CoT & 1{,}077 & 942 & 87.47 & 940 & 38\\
\midrule
ChartQA  & InternVL & FF & 1{,}077 & 984 & 91.36 & 982 & 28\\*
 & & SRT & 1{,}077 & 982 & 91.18 & 981 & 28\\*
 & & CoT & 1{,}077 & 983 & 91.27 & 982 & 29\\
\midrule
ChartQA  & MiniCPM & FF & 1{,}077 & 870 & 80.78 & 869 & 55\\*
 & & SRT & 1{,}077 & 870 & 80.78 & 868 & 55\\*
 & & CoT & 1{,}077 & 862 & 80.04 & 862 & 58\\
\midrule
Mixture & Qwen2.5 & FF & 1{,}306 & 1{,}123 & 85.99 & 1{,}079 & 97\\*
 & & SRT & 1{,}306 & 1{,}120 & 85.76 & 1{,}063 & 97\\*
 & & CoT & 1{,}306 & 1{,}097 & 84.00 & 1{,}090 & 128\\
\midrule
Mixture  & Qwen3.5 & FF & 1{,}306 & 1{,}149 & 87.98 & 1{,}101 & 133\\*
 & & SRT & 1{,}306 & 1{,}164 & 89.13 & 1{,}142 & 85\\*
 & & CoT & 1{,}306 & 1{,}139 & 87.21 & 1{,}116 & 118\\
\midrule
Mixture  & Molmo & FF & 1{,}306 & 966 & 73.97 & 921 & 171\\*
 & & SRT & 1{,}306 & 914 & 69.98 & 871 & 194\\*
 & & CoT & 1{,}306 & 930 & 71.21 & 930 & 195\\
\end{longtable}
\endgroup

Table~\ref{tab:s-contingency} shows that correct states and correct answers can diverge in either direction. With a correct state, Qwen2.5 with SRT misanswers 7 of 962 pair-state ChartQA questions but 48 of 245 multi-step arithmetic questions at seed 42. Conversely, the model answers 72 of 86 ranking questions correctly despite a wrong state. These counts distinguish errors in writing visual facts from errors in applying the requested operation.

\FloatBarrier
\begingroup
\small
\setlength{\tabcolsep}{4.5pt}
\renewcommand{\arraystretch}{0.91}
\setlength{\LTpre}{6pt}
\setlength{\LTpost}{6pt}
\begin{longtable}{@{\extracolsep{\fill}}llrrr@{}}
\caption{\textbf{State--answer contingency counts under SRT.} Each entry gives count / denominator within the indicated generated-state outcome.}
\label{tab:s-contingency}\\
\toprule
Setting & Model & Seed & Wrong state, answer correct & Correct state, answer wrong\\
\midrule
\endfirsthead
\multicolumn{5}{@{}l}{\textbf{Table \thetable.} Continued from the preceding page.}\\[3pt]
\toprule
Setting & Model & Seed & Wrong state, answer correct & Correct state, answer wrong\\
\midrule
\endhead
\midrule
\multicolumn{5}{r@{}}{\footnotesize Continued on the next page.}\\
\endfoot
\bottomrule
\endlastfoot
ChartQA & Qwen2.5 & 13 & 36 / 120 & 10 / 957\\*
 & & 42 & 36 / 115 & 7 / 962\\*
 & & 87 & 33 / 117 & 3 / 960\\
\midrule
ChartQA  & Qwen3.5 & 42 & 28 / 82 & 0 / 995\\
\midrule
ChartQA  & Molmo & 13 & 38 / 125 & 1 / 951\\*
 & & 42 & 37 / 122 & 3 / 955\\*
 & & 87 & 40 / 132 & 1 / 939\\
\midrule
ChartQA  & InternVL & 13 & 27 / 91 & 2 / 986\\*
 & & 42 & 28 / 95 & 1 / 982\\*
 & & 87 & 25 / 82 & 3 / 995\\
\midrule
ChartQA  & MiniCPM & 13 & 46 / 190 & 2 / 887\\*
 & & 42 & 55 / 207 & 2 / 870\\*
 & & 87 & 58 / 209 & 2 / 868\\
\midrule
CLEVR & Qwen2.5 & 42 & 517 / 2{,}072 & 0 / 31{,}651\\
\midrule
Multi-step arithmetic & Qwen2.5 & 42 & 0 / 46 & 48 / 245\\
Conditional lookup & Qwen2.5 & 42 & 19 / 44 & 7 / 171\\
Relational count ranking & Qwen2.5 & 42 & 72 / 86 & 0 / 314\\
Color-set operations & Qwen2.5 & 42 & 6 / 10 & 2 / 390\\
\end{longtable}
\endgroup

\FloatBarrier
\section{Source-Bound Verification with Parallax}
\label{sec:s-arbitration}

\subsection{Training and Inference Protocol}
We freeze both generation routes of Qwen3.5-9B~\cite{qwenteam2026qwen35} and train only a LoRA selector~\cite{hu2022lora}. Both routes are evaluated on all 28,299 official ChartQA training questions. A question supplies a directional training target only when both final answers are valid, their normalized forms differ, and exactly one is correct under the official binary scorer. The 4,433 eligible questions include 3,515 whose target is to keep Direct and 918 whose target is to switch to the challenger. Presenting both option orders yields 8,866 training records. We train for one epoch with an effective batch size of 8 and seed 42, using LoRA rank 16, scaling parameter 32, dropout 0.05, and AdamW~\cite{loshchilov2019decoupled} with a learning rate of $10^{-4}$. The loss is unweighted full-vocabulary cross-entropy on the correct option token and the assistant EOS token.

At inference, the gate uses no reference answer. The selector receives the image, question, and candidate answers as Options A and B (Section~\ref{sec:s-parallax-prompts}). The source header and withheld-draft marker follow Challenger when the options swap. We normalize the A/B scores, average the Challenger probability across the two orders, and switch to Challenger only if the gate admits the pair and this probability is at least 0.5. Without the header, the source and draft block is omitted; the full-draft variant exposes the draft under the same header. All three variants share the initialization, random-number generator state, training pairs, presentation order, and optimization settings. Direct and Challenger can run in parallel because neither takes the other's output. All three components use the same image and frozen vision encoder; only gated disagreements ($27.7\%$ of original ChartQA test questions) require the selector's two next-token scoring passes, with no autoregressive decoding.

\FloatBarrier
\subsection{Selector Ablations and Reasoning Baselines}
On all 2,500 official ChartQA test questions (Table~\ref{tab:s-natural-arbitration}), Parallax captures 125 of 129 valid repairs with six regressions and wins 26 questions against two losses versus the symmetric selector ($90.48\%$). The source-hidden control ($90.72\%$) retains the wrapper, isolating the contribution of the route label, while exposing the full draft ($91.32\%$) does not improve on Parallax ($91.44\%$). Correct answers can accompany inaccurate visual facts (Section~\ref{sec:s-state-contingency}), so withholding the draft lets the selector check the answers without relying on those claims. Removing or mismatching the image raises regressions from six to 37 or 48, respectively (Table~7 of the main paper), showing that the image helps prevent erroneous switches. Untrained scoring ($86.16\%$) and Zero-shot judge ($88.44\%$) use the frozen VLM and the withheld- and full-draft templates, respectively, with the latter restricted to the two existing answers~\cite{zheng2023judging}.

\begin{table}[!htbp]
\centering
\small
\renewcommand{\arraystretch}{0.9}
\caption{\textbf{Comparison of Parallax selector variants on ChartQA ($N{=}2{,}500$).} All selector variants evaluate the same candidate answers on the 693 gated disagreements.}
\label{tab:s-natural-arbitration}
\setlength{\tabcolsep}{6pt}
\begin{tabular}{@{}lrr@{}}
\toprule
Condition & Correct & Acc. (\%) \\
\midrule
Direct & 2,167 & 86.68 \\
Challenger alone & 1,785 & 71.40 \\
Without source header & 2,262 & 90.48 \\
Source-hidden control & 2,268 & 90.72 \\
With full reasoning draft & 2,283 & 91.32 \\
Parallax & \textbf{2,286} & \textbf{91.44} \\
\bottomrule
\end{tabular}
\end{table}

\begin{table}[!htbp]
\centering
\small
\setlength{\tabcolsep}{4pt}
\renewcommand{\arraystretch}{0.9}
\caption{\textbf{Reasoning baselines vs.\ Parallax on natural ChartQA ($N{=}2{,}500$).} Prompts, decoding modes, and token caps follow each system's configuration.}
\label{tab:s-natural-cot-profiles}
\begin{tabular}{@{}lrr@{}}
\toprule
System & Correct & Acc. (\%) \\
\midrule
Direct & 2,167 & 86.68 \\
Step-by-step CoT & 1,966 & 78.64 \\
Challenger CoT & 1,785 & 71.40 \\
Zero-shot CoT prompt (512 cap) & 1,571 & 62.84 \\
Zero-shot CoT prompt (4,096 cap) & 1,666 & 66.64 \\
Thinking mode (8,192 cap) & 1,878 & 75.12 \\
Parallax & \textbf{2,286} & \textbf{91.44} \\
\bottomrule
\end{tabular}
\end{table}

In Table~\ref{tab:s-natural-cot-profiles}, raising the zero-shot CoT cap from 512 to 4,096 tokens improves accuracy from 62.84\% to 66.64\%. Native thinking reaches 75.12\% with an 8,192-token cap and averages 1,012.66 output tokens. Parallax exceeds each standalone configuration.

\begin{table}[!htbp]
\centering
\small
\setlength{\tabcolsep}{3.5pt}
\renewcommand{\arraystretch}{0.9}
\caption{\textbf{Selection on Molmo candidates on original ChartQA ($N{=}2{,}500$).} Repairs and regressions are measured against Molmo Direct, where reported. The transferred selector is trained on Qwen3.5 candidates and applied to Molmo candidates without retraining.}
\label{tab:s-molmo-transfer}
\begin{tabular*}{\textwidth}{@{\extracolsep{\fill}}lrrrr@{}}
\toprule
System on Molmo candidates & Correct & Acc. (\%) & Repairs & Regr. \\
\midrule
Molmo Direct & 1,450 & 58.00 & -- & -- \\
Molmo Challenger & 1,288 & 51.52 & 563 & 725 \\
Molmo Step-by-step CoT & 1,604 & 64.16 & -- & -- \\
Untrained Molmo selector & 1,525 & 61.00 & -- & -- \\
Parallax (Molmo selector) & 1,980 & 79.20 & 538 & 8 \\
Parallax (Qwen3.5 selector) & \textbf{1,999} & \textbf{79.96} & \textbf{560} & 11 \\
\bottomrule
\end{tabular*}
\end{table}

\noindent\textbf{Native and transferred selectors on Molmo.}
Both option orders of 12,272 eligible Molmo training questions yield 24,544 records. We train only the selector with seed 42, effective batch size eight, and the same option-and-EOS objective. Both selectors score the same 1,662 gated disagreements.

In Table~\ref{tab:s-molmo-transfer}, the trained Molmo selector reaches 79.20\%, gaining 18.20 points over the untrained selector (61.00\%; 95\% image-cluster bootstrap CI $[16.58,19.85]$ for the gain) and 21.20 points over Molmo Direct, with 538 repairs and eight regressions. Challenger alone is correct on 20.7\% (918/4,433) of eligible Qwen3.5 training questions and on 43.7\% (563/1,288) of single-winner Molmo test questions before gating. Without retraining, the Qwen3.5 selector reaches 79.96\% and captures 99.5\% (560/563) of valid repairs with 11 regressions. This transfer is consistent with question-specific selection rather than a fixed route preference.

\FloatBarrier
\subsection{Transfer to StateLens Questions}
We match the 1,077 StateLens ChartQA questions to the original Qwen3.5 diagnosis by question and image identifiers. Without retraining, Parallax answers 983 correctly (91.27\%), with 288 repairs out of 291 valid opportunities and 12 regressions. It recovers 256 of the 259 repairable cases among 268 diagnosed composition failures.

\FloatBarrier
\newenvironment{suppqualfigure}{%
  \par\addvspace{6pt}\noindent\begin{minipage}{\textwidth}\centering
  \captionsetup{hypcap=false,skip=5pt}%
}{\end{minipage}\par\addvspace{6pt}}

\section{Additional Qualitative Examples}
\label{sec:s-examples}
Figure~\ref{fig:s-behavior} contrasts state and calculation errors; Figures~\ref{fig:s-clevr-pair}--\ref{fig:s-clevr-rel} give further AO and SRT examples.

\begin{suppqualfigure}
\centering
\includegraphics[width=0.69\textwidth,trim={0pt 34pt 0pt 1pt},clip]{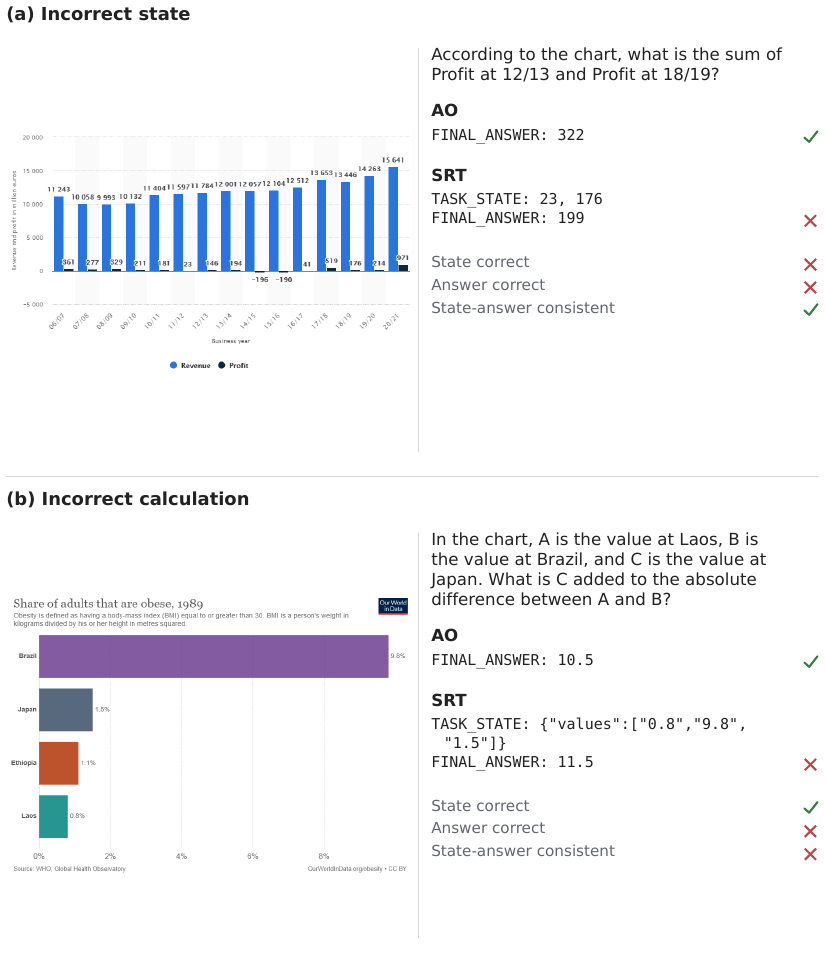}
\captionof{figure}{\textbf{Distinction between state accuracy and state--answer consistency.} (a) On ChartQA, Qwen2.5 sums its two written values consistently, but one extracted value is wrong. (b) On multi-step arithmetic, Molmo extracts all three values accurately but errs in the two-step calculation.}
\label{fig:s-behavior}
\end{suppqualfigure}
\begin{suppqualfigure}
\centering
\includegraphics[width=0.75\textwidth,trim={0pt 3pt 0pt 3pt},clip]{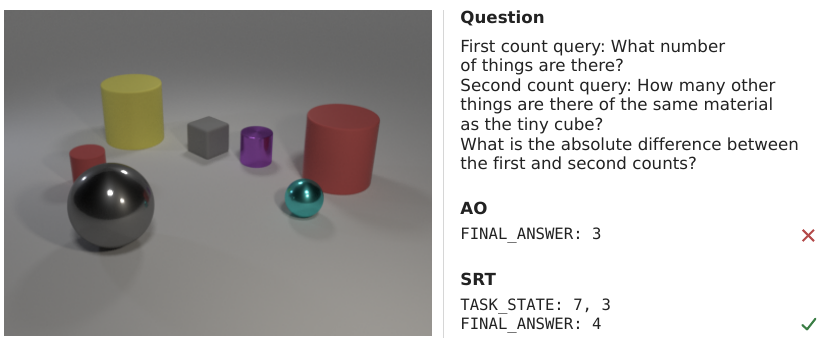}
\captionof{figure}{\textbf{Pair-state CLEVR example.} Qwen2.5 with AO answers the difference question incorrectly. With SRT, it writes both verified counts before computing their difference.}
\label{fig:s-clevr-pair}
\end{suppqualfigure}

\begin{suppqualfigure}
\centering
\includegraphics[width=0.78\textwidth,trim={0pt 14pt 0pt 14pt},clip]{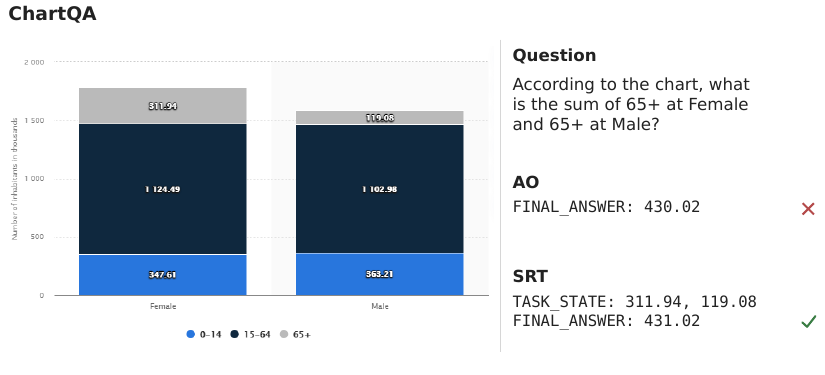}
\captionof{figure}{\textbf{Pair-state ChartQA example.} Qwen2.5 with AO misses the decimal sum. With SRT, it writes the two bar values in question order and outputs the exact sum.}
\label{fig:s-chartqa-pair}
\end{suppqualfigure}

\begin{suppqualfigure}
\centering
\includegraphics[width=0.68\textwidth,trim={0pt 5pt 0pt 4pt},clip]{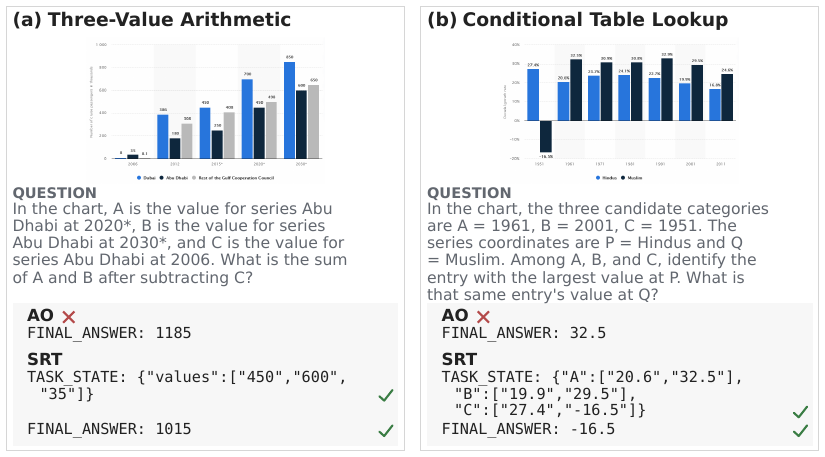}
\captionof{figure}{\textbf{Three-value arithmetic and conditional lookup.} Qwen2.5 with SRT writes the three operands or the $3\times2$ table before answering correctly, whereas AO errs on both questions.}
\label{fig:s-hard-num}
\end{suppqualfigure}
\begin{suppqualfigure}
\centering
\includegraphics[width=0.64\textwidth,trim={0pt 0pt 0pt 4pt},clip]{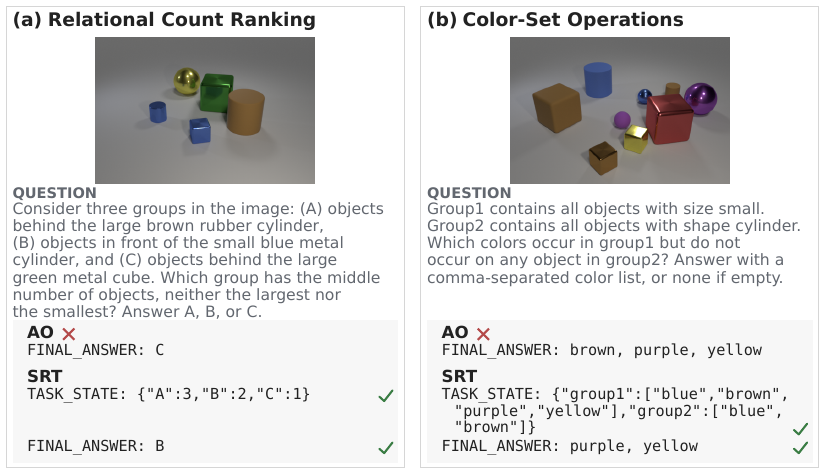}
\captionof{figure}{\textbf{Relational count ranking and color-set difference.} Qwen2.5 with SRT writes the three counts before selecting the middle group and both color sets before taking their difference.}
\label{fig:s-hard-struct}
\end{suppqualfigure}

\begin{suppqualfigure}
\centering
\includegraphics[width=0.52\textwidth,trim={0pt 20pt 0pt 1pt},clip]{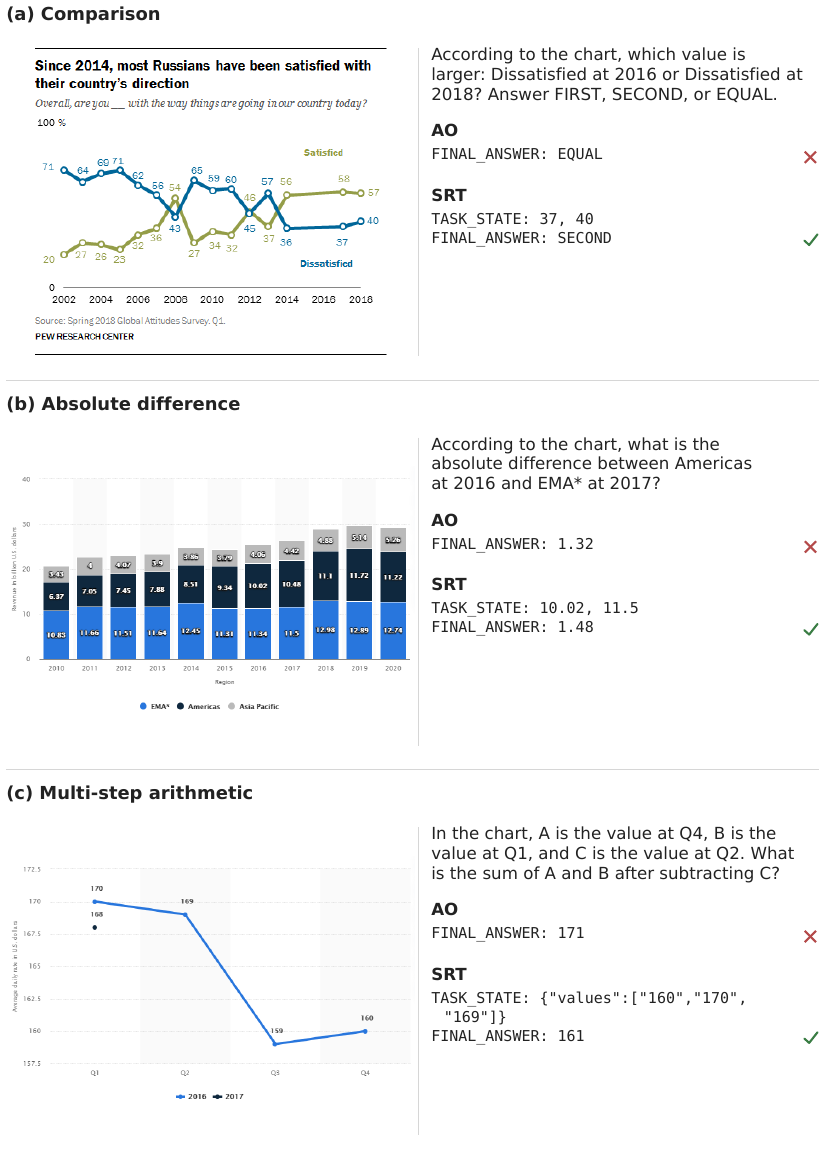}
\captionof{figure}{\textbf{Diverse chart styles and operations.} (a) Line-chart comparison (Qwen2.5). (b) Stacked-bar absolute difference (Qwen2.5). (c) Multi-series line chart with three-value arithmetic on the 2016 series (Molmo).}
\label{fig:s-diversity}
\end{suppqualfigure}
\begin{suppqualfigure}
\centering
\includegraphics[width=0.73\textwidth,trim={0pt 25pt 0pt 1pt},clip]{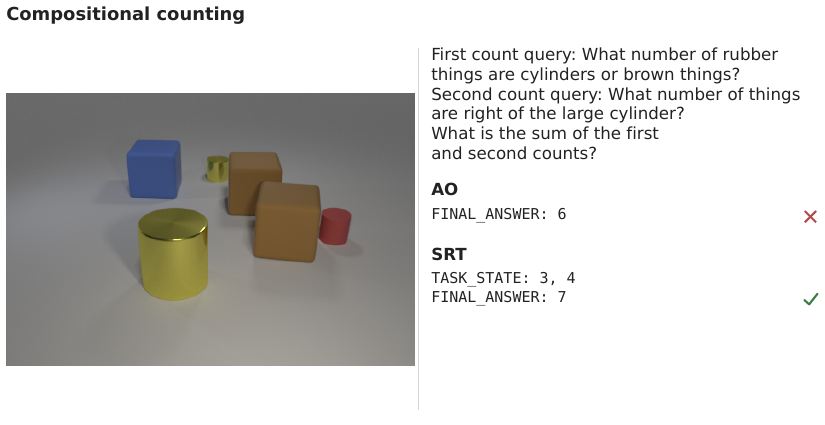}
\captionof{figure}{\textbf{Compositional CLEVR counting.} Of the two count queries, the first combines material filtering with a union of categories, and the second counts objects to the right of a reference cylinder.}
\label{fig:s-clevr-rel}
\end{suppqualfigure}

\end{document}